\documentclass[10pt,twocolumn,letterpaper]{article}
\usepackage[pagenumbers]{cvpr}      
\definecolor{cvprblue}{rgb}{0.21,0.49,0.74}
\usepackage[pagebackref,breaklinks,colorlinks,allcolors=cvprblue]{hyperref}

\def\paperID{7196} 
\def\confName{CVPR}
\def\confYear{2026}

\title{Gradient-Driven Natural Selection for Compact 3D Gaussian Splatting}

\author{ Xiaobin Deng \quad Qiuli Yu \quad  Changyu Diao\footnotemark[1] \quad Min Li \quad Duanqing Xu\footnotemark[1]\\
Zhejiang University\\}

\begin{document}
\twocolumn[{
    \maketitle
	\vspace{-2em}
	\includegraphics[width=1.\linewidth]{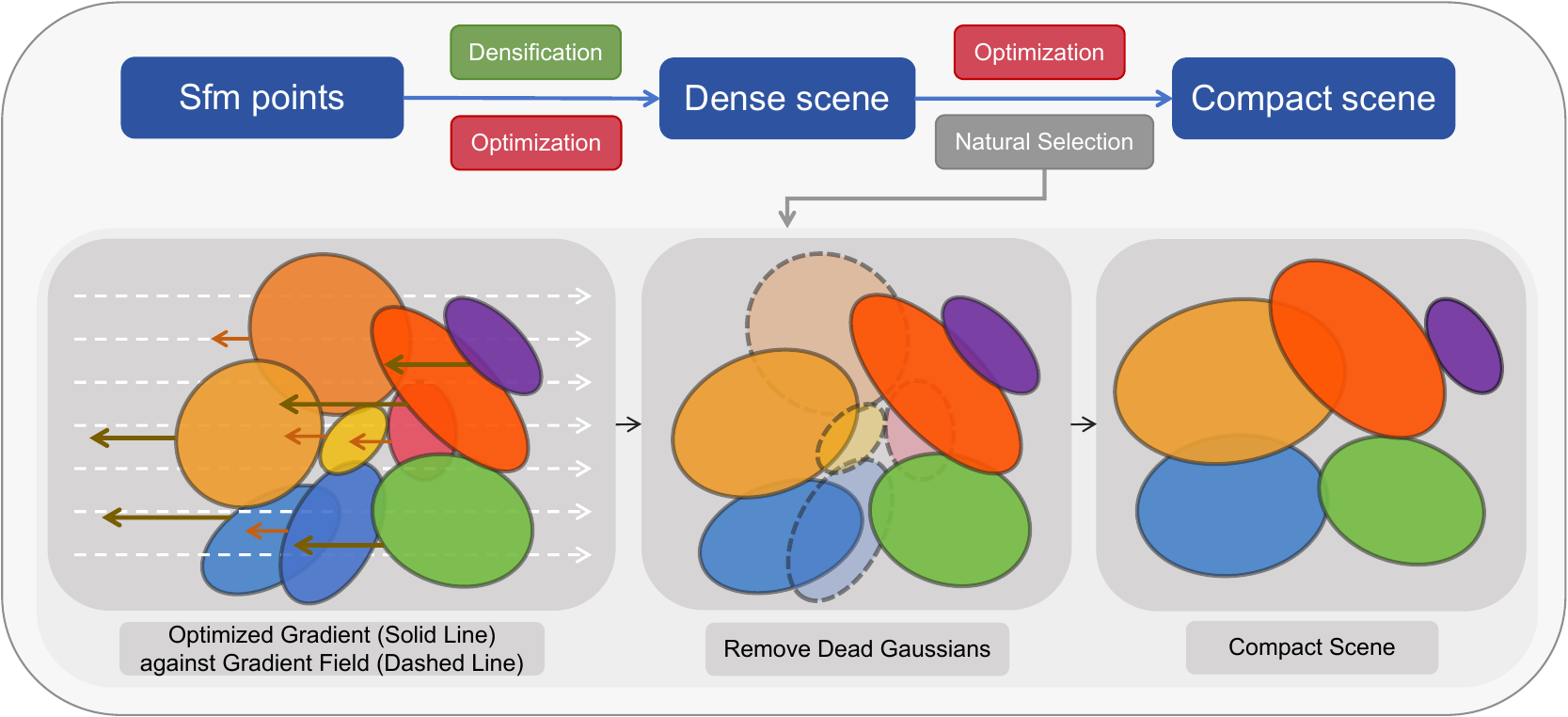}
    \vspace{-20pt}
    \captionof{figure}{Our method's pipeline consists of two parts. 
    The upper section shows the full workflow: starting from sparse SfM point clouds, we densify and optimize to obtain a high-quality dense scene, which is then refined via a natural selection mechanism to produce a compact, high-fidelity representation. 
    The lower section details this natural selection framework: a globally consistent regularization gradient field (dashed lines) is applied to all Gaussian opacities, guiding optimization gradients (solid lines) to identify and prune Gaussians whose opacity falls below a survival threshold. 
    Those Gaussians with smaller optimized gradients will gradually decay in opacity until reaching the death threshold, after which they are permanently removed.
    The result is a high-quality, compact scene.}
	\vspace{1.0em}
    \label{fig:intro}
}]
{
\renewcommand{\thefootnote}{\fnsymbol{footnote}}
\footnotetext[1]{Corresponding authors.}
}
\begin{abstract}
3DGS employs a large number of Gaussian primitives to fit scenes, resulting in substantial storage and computational overhead. 
Existing pruning methods rely on manually designed criteria or introduce additional learnable parameters, yielding suboptimal results. 
To address this, we propose an natural selection inspired pruning framework that models survival pressure as a regularization gradient field applied to opacity, allowing the optimization gradients—driven by the goal of maximizing rendering quality—to autonomously determine which Gaussians to retain or prune. 
This process is fully learnable and requires no human intervention. 
We further introduce an opacity decay technique with a finite opacity prior, which accelerates the selection process without compromising pruning effectiveness.
Compared to 3DGS, our method achieves over 0.6 dB PSNR gain under 15\% budgets, establishing state-of-the-art performance for compact 3DGS.
Project page \url{https://xiaobin2001.github.io/GNS-web} .
\end{abstract}    
\section{Introduction}
\label{sec:intro}

Novel view synthesis is a fundamental yet challenging task in computer vision, with the core objective of reconstructing scene images from novel viewpoints given a limited set of input views.
This technology demonstrates significant value in critical domains such as virtual reality (VR), digital twins, and autonomous driving. 
Neural Radiance Fields (NeRF) \cite{mildenhall2021nerf}, based on implicit neural representations, achieve a breakthrough in high-fidelity view synthesis from sparse input views by parameterizing the light transport function using Multilayer Perceptron (MLP). 
However, this method struggles to achieve real-time rendering. 
In contrast, 3D Gaussian Splatting (3DGS) \cite{kerbl20233d}, as a reconstruction technique utilizing explicit representation, simultaneously ensures both rendering quality and real-time performance. 
Initialized from sparse point clouds obtained through SfM (Structure from Motion) \cite{schonberger2016structure}, 3DGS models the scene as a collection of 3D Gaussian primitives with optimizable parameters.
Each primitive is characterized by its center position $\mathbf{\mu}$, covariance matrix $\mathbf{\Sigma}$, opacity $\alpha$, and color attributes encoded via spherical harmonics.

3DGS typically employs millions of Gaussians to fit a scene for high-quality rendering. 
However, the large number of Gaussian ellipsoids significantly increases rendering and storage costs, hindering the adoption of downstream applications. 
Existing compact 3DGS variants reduce the number of Gaussians through manually designed pruning criteria or by introducing additional learnable masks.
Inspired by natural selection, we propose a learnable pruning technique that requires neither manually designed pruning criteria nor additional parameters, thereby enhancing the performance of compact 3DGS.

\begin{figure}[t]
\centering
\includegraphics[width=0.36\textwidth]{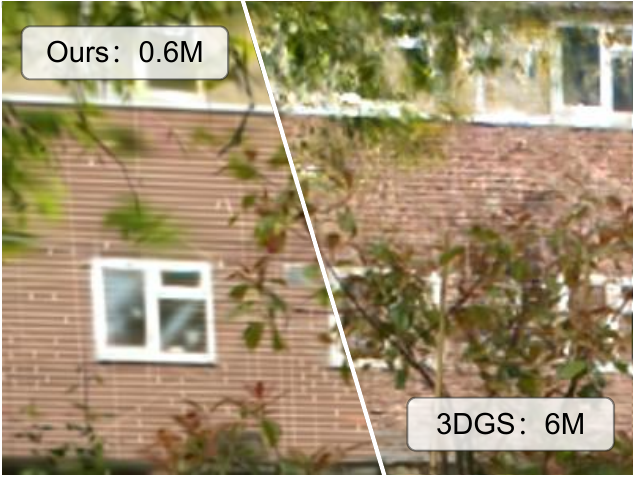}
\caption{Qualitative comparison between our method and 3DGS on the garden scene}
\label{fig:intro2}
\end{figure}

Figure~\ref{fig:intro} illustrates the overall pipeline of our proposed method.
Inspired by natural selection, where environmental pressures screen for the fittest genes, we model a sparsity metric as a uniform survival pressure applied to all Gaussians, gradually reducing their opacity. 
Concurrently, optimization gradients aimed at maximizing rendering quality counteract the negative gradients induced by this pressure. 
This process favors Gaussians that contribute more significantly to rendering quality, as those receiving stronger optimization gradients survive longer under high survival pressure. 
To confine the natural selection process within a limited number of training iterations, we design a minimal-prior opacity decay technique to simulate environmental pressure. 
Under the same training iterations, our method uses only 15\% of the budget and one-third of the training time compared to the original 3DGS, while achieving a 0.6 dB improvement in PSNR.
Figure~\ref{fig:intro2} compares our method with the original 3DGS.

In summary, our contributions are as follows:
\begin{itemize}
  
    \item We propose a learnable pruning technique that requires neither manually designed pruning criteria nor additional parameters, yielding highly competitive performance.
    \item We design a opacity decay technique with finite that accelerates the pruning process without compromising the resulting quality.
    \item Our method achieves state-of-the-art performance in compact 3DGS and is highly portable.

\end{itemize}
\section{Related Works}
\label{sec:relat}

\textbf{Quality Optimization of 3DGS: }
3DGS already achieves a balance between rendering quality and real-time performance, yet there remains room for optimization. 
AbsGS \cite{ye2024absgs} addresses the inherent reconstruction blur in 3DGS by introducing an absolute gradient-based densification evaluation criterion. 
TamingGS \cite{mallick2024taming} incorporates multiple evaluation metrics for the densification process and optimizes the rendering kernel of 3DGS, significantly improving training speed. 
Improved-GS \cite{deng2025improving} reconstructs the densification process of 3DGS from three perspectives, significantly enhancing rendering quality.
ScaffoldGS \cite{lu2024scaffold} combines implicit and explicit representations by using neural networks to implicitly store the parameters of Gaussian primitives. 
Mip-Splatting \cite{yu2024mip} introduces a 3D smoothing filter and a 2D mipmap filter to mitigate aliasing artifacts that may occur during magnification in 3DGS. 
GaussianPro \cite{cheng2024gaussianpro} utilizes optimized depth and normal maps to initialize new Gaussians via reprojection. It further enhances geometric reconstruction through planar regularization. 
2DGS \cite{huang20242d} employs 2D Gaussian primitives to represent scenes, improving the application of Gaussian Splatting in geometric reconstruction. 
Spec-Gaussian \cite{yang2024spec} adopts an anisotropic spherical Gaussian appearance field for Gaussian color modeling, significantly enhancing rendering quality in complex scenes with specular reflections and anisotropic surfaces.

\textbf{Compact 3DGS: }
The approaches to achieving compact 3DGS primarily fall into two categories. 
One builds upon the neural representation introduced by ScaffoldGS, utilizing octrees \cite{ren2024octree} or hash grids \cite{chen2024hac} for compression. This approach relies on specific neural representations and cannot be generalized to standard 3DGS. The other approach involves pruning techniques to eliminate redundant Gaussians, thereby reducing the total number of Gaussians in the scene to achieve compactness.

Compact3DGS \cite{lee2024compact} aims to reduce both rendering and storage overhead in 3DGS. It proposes a learnable masking strategy to decrease the number of Gaussians, while further compressing storage through neural color representations and vector quantization.
LightGaussian \cite{fan2024lightgaussian} employs a global rendering weight combined with Gaussian volume to assist in pruning. LP-3DGS introduces the Gumbel-Sigmoid activation function into 3DGS to replace the Sigmoid function for masking or pruning scoring. Gumbel-Sigmoid pushes values closer to 0 or 1, providing a good approximation for binary masks.
Compact3DGS  uses the straight-through estimator to address the non-differentiability issue in the binarization process, whereas MaskGS \cite{liu2025maskgaussian} directly employs probabilistic masks and derives gradients for mask parameters, achieving better mask pruning results. MaskGS also utilizes Gumbel-Sigmoid for activating mask parameters.
Mini-Splatting \cite{fang2024mini} introduces depth reinitialization to address the uneven spatial distribution of Gaussians and uses maximum rendering contribution area and global rendering weight as criteria for pruning.
GaussianSPA \cite{zhang2025gaussianspa} employs the Alternating Direction Method of Multipliers (ADMM) to gradually attenuate the opacity of Gaussians scheduled for pruning to zero during optimization. This replaces explicit pruning with a smooth optimization process, contributing to improved rendering quality.
Unlike previous methods, we avoid manual pruning rules and extra parameters.

Based on the different technical approaches adopted, we select the following works as comparative benchmarks: Compact3DGS, as a representative of binary masking; Mini-Splatting, as a representative of scene reorganization for sparsification; MaskGS, as a representative of probabilistic masking; and GaussianSPA, as a representative of progressive sparsification.
LP-3DGS \cite{zhang2024lp} only introduces a new activation function, which is also used in MaskGS, and is therefore not included in the comparison. The rendering importance pruning employed by LightGaussian will be compared in the ablation experiments.
\section{Methods}
\setlength{\tabcolsep}{2pt}
\label{sec:methods}

\subsection{Preliminaries}
In 3DGS framework, a scene is represented by a collection of anisotropic 3D Gaussian primitives:
\begin{equation}
    G(x)=\exp{\left( -\frac{1}{2} (x)^T \Sigma^{-1} (x) \right)},
\end{equation}
where $x$ denotes the offset from the Gaussian’s mean position, and $\Sigma$ is its 3D covariance matrix.  
To guarantee that $\Sigma$ remains positive semi-definite, 3DGS expresses it through a decomposition involving a rotation matrix $R$ and a scaling matrix $S$:
\begin{equation}
    \Sigma = R S S^T R^T.
\end{equation}
Here, the scaling matrix $S$ is parameterized by a 3D vector $s$, and the rotation matrix $R$ is derived from a unit quaternion $q$.  
When rendering an image from a given camera viewpoint, the final color of a pixel $p$ is computed by alpha-compositing $N$ Gaussians $\{ G_i \mid i = 1, \dots, N \}$ that project onto $p$, ordered from front to back, according to:
\begin{equation}
	\label{eq:front-to-back}
	C = \sum_{i=1}^{N}
	c_{i}\alpha_{i}
	\prod_{j=1}^{i-1}(1-\alpha_{j}),
\end{equation}
where $\alpha_i$ is the opacity contribution of $G_i$ at pixel $p$ (obtained by evaluating the projected Gaussian and scaling by its intrinsic opacity), and $c_i$ is the color of $G_i$, encoded using spherical harmonics (SH) coefficients.

\subsection{Compact 3DGS Objective}
\label{sec:objective}

\textbf{Background and Challenges:}
Under a constrained Gaussian budget, the core objective of compact 3DGS remains to achieve high-fidelity rendering. 
Traditional 3DGS densification strategies identify blurry regions and insert additional Gaussians, which are inherently unconstrained local operations. 
However, when the total number of Gaussians (budget) is limited, global balance of rendering quality becomes essential. 
Simply restricting densification growth is therefore insufficient to meet the requirement of compact 3DGS (see Sec.~\ref{sec:quantitative_omparison}).

Current approaches typically adopt a two-stage strategy: first, generate a high-quality scene representation through unconstrained densification; second, perform pruning to balance quality and budget.

\textbf{Bi-Objective Decomposition:}
In 3DGS, scene optimization relies on back-propagation, where each Gaussian only receives gradient signals from pixels it contributes to. 
Consequently, the retained subset of Gaussians after pruning, denoted as \( G_C \), fundamentally determines the achievable rendering upper bound. 
Once a Gaussian representing unique local details is mistakenly pruned, its contribution is almost unrecoverable.

Based on this, we decompose the overall objective of compact 3DGS into two sub-goals:

\begin{enumerate}
    \item Selecting a superior retained subset \( G_{C} \) to maximize the potential quality upper bound;
    \item Ensuring stable convergence of the pruning-optimization process within a limited number of iterations.
\end{enumerate}

The following sections address these two goals respectively. Sec.~\ref{sec:NS} introduces a natural selection–based adaptive pruning framework for achieving Goal 1, while Sec.~\ref{sec:RGF} presents a regularization gradient field with finite priors to accelerate convergence (Goal 2).

\subsection{Adaptive Pruning via Natural Selection}
\label{sec:NS}

To fulfill Goal 1—selecting a superior subset \( G_{C} \)—we draw inspiration from natural selection in biological evolution (excluding genetic inheritance and mutation) and propose an adaptive pruning framework. This framework simulates the competition between environmental pressure and individual fitness, progressively eliminating low-contribution Gaussians and retaining high-importance ones to maximize rendering quality under a strict budget.

\textbf{Problem Formulation:}
Let the complete Gaussian set be \( G = \{ g_i \}_{i=1}^N \), where each Gaussian \( g_i \) is parameterized as \( \theta_i = (\alpha_i, \mu_i, \Sigma_i, c_i) \) (opacity, mean, covariance, and color). The pruning objective is formulated as:

\begin{equation}
    \min_{G_{C} \subset G} \mathcal{L}_{\text{render}}(G_{C})
    \quad \text{s.t.} \quad |G_{C}| \le B,
\end{equation}

where \( \mathcal{L}_{\text{render}} \) denotes the rendering loss and \( B \) is the Gaussian budget.

\textbf{Natural Selection Mechanism:}
We model the biological analogy through the following definitions:
\begin{itemize}
    \item Vitality: the opacity \( \alpha_i \in [0,1] \) represents the existence strength of each Gaussian. This is because opacity directly determines the rendering contribution of each primitive.
    \item Environmental Pressure: a globally consistent regularization gradient field \( \mathcal{L}_{\text{reg}}(\alpha) \) applied to all Gaussians (see Sec.~\ref{sec:RGF} for details).
    \item Fitness: the rendering gradient \( \nabla_{\alpha_i} \mathcal{L}_{\text{render}} \) received by Gaussian \( g_i \), representing its dynamic importance.
\end{itemize}
The overall loss is defined as:
\begin{equation}
    \mathcal{L} = \mathcal{L}_{\text{render}}(\Theta) + \mathcal{L}_{\text{reg}}(\alpha),
\end{equation}
and the net gradient on the opacity of each Gaussian is:
\begin{equation}
    \nabla_{\alpha_i}^{\text{net}}
    = \nabla_{\alpha_i} \mathcal{L}_{\text{render}} + \nabla_{\alpha} \mathcal{L}_{\text{reg}}.
\end{equation}
Here, the first term is Gaussian-specific, while the second term is a global constant. 
In practice, the regularization gradient is applied every \( N = 50 \) rendering iterations. 
This allows fitness signals to accumulate from multiple training views, producing more stable survival decisions before applying environmental pressure.

If the rendering gradient \( \nabla_{\alpha_i} \mathcal{L}_{\text{render}} \) is opposite in direction to the regularization gradient \( \nabla_{\alpha} \mathcal{L}_{\text{reg}} \), increasing opacity improves rendering quality, and the Gaussian tends to survive. 
Conversely, when both gradients are aligned, the Gaussian contributes negatively to the rendering and is rapidly suppressed. Since the regularization gradient magnitude is globally constant, the relative differences between rendering gradients dominate the competition—only high-fitness Gaussians maintain strong vitality and survive.

At the end of densification, when the scene has largely converged, the natural selection process is activated. 

\textbf{Selection Process and Survival Criterion:}

During optimization, a Gaussian is permanently removed if its opacity falls below a survival threshold: \(  \alpha_i < \tau. \)
We empirically set $\tau = 0.001$, smaller than the $0.005$ threshold used in 3DGS, as Gaussians below this level contribute negligibly to rendering.

Strong environmental pressure gradually removes weak Gaussians until \( |G_{C}| \le B \). The resulting subset \( G_{C} \) thus consists solely of high-fitness, high-contribution primitives.

Importantly, this framework requires no pre-defined importance score; instead, global competition among gradients naturally yields an optimal subset.

\textbf{Key Innovations:}
Prior works regularize opacity in two common ways, both fundamentally different from ours:

\begin{enumerate}
    \item Weak Regularization: introduces low-intensity penalties to reduce average opacity and encourage color aggregation from multiple primitives.
    \item Auxiliary Pruning Regularization: applies regularization as a post-processing tool for numerical stability, but it does not participate in pruning decisions.
\end{enumerate}
In contrast, our method applies a strong, globally consistent regularization gradient that explicitly competes with the rendering gradient. This introduces a novel gradient-competition-driven pruning mechanism, enabling:

\begin{enumerate}
    \item Automatic Selection: weak Gaussians are eliminated via gradient competition;
    \item Smooth Optimization: opacity decays continuously, avoiding instability from discrete pruning.
\end{enumerate}

As shown in Section~\ref{sec:quantitative_omparison}, this mechanism consistently outperforms hand-designed pruning heuristics and learnable-mask approaches across multiple datasets.

\begin{figure}[t]
    \centering
    \includegraphics[width=0.47\textwidth]{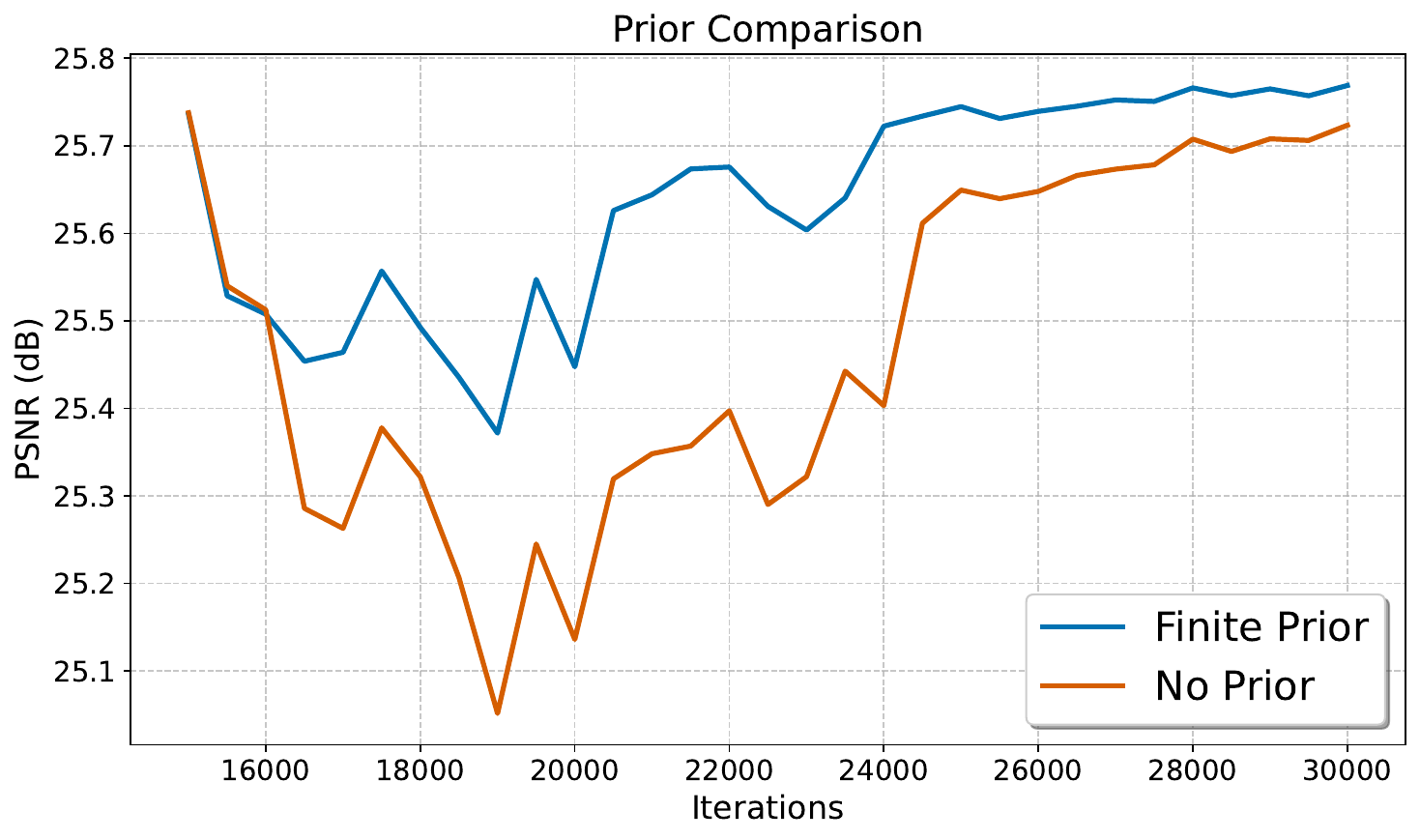}
    \caption{The figure illustrating the optimization speed improvement of finite prior over no prior. }
    \label{fig:prior}
\end{figure}

\subsection{Gradient Field with Finite Priors}
\label{sec:RGF}

\textbf{Introduction of Finite Priors:}
In the natural selection mechanism, the gradient of the optimization loss with respect to opacity parameters is used to counteract the environmental pressure, while other parameters continue to be optimized to minimize the rendering loss. 
To achieve faster optimization convergence (Goal 2 in Section~\ref{sec:objective}), it is essential to design an appropriate environmental pressure that ensures the natural selection process does not hinder the optimization of opacity parameters for surviving Gaussians.

\begin{figure*}[t]

\centering
\begin{minipage}[b]{0.32\textwidth}
\centering
\includegraphics[width=\textwidth]{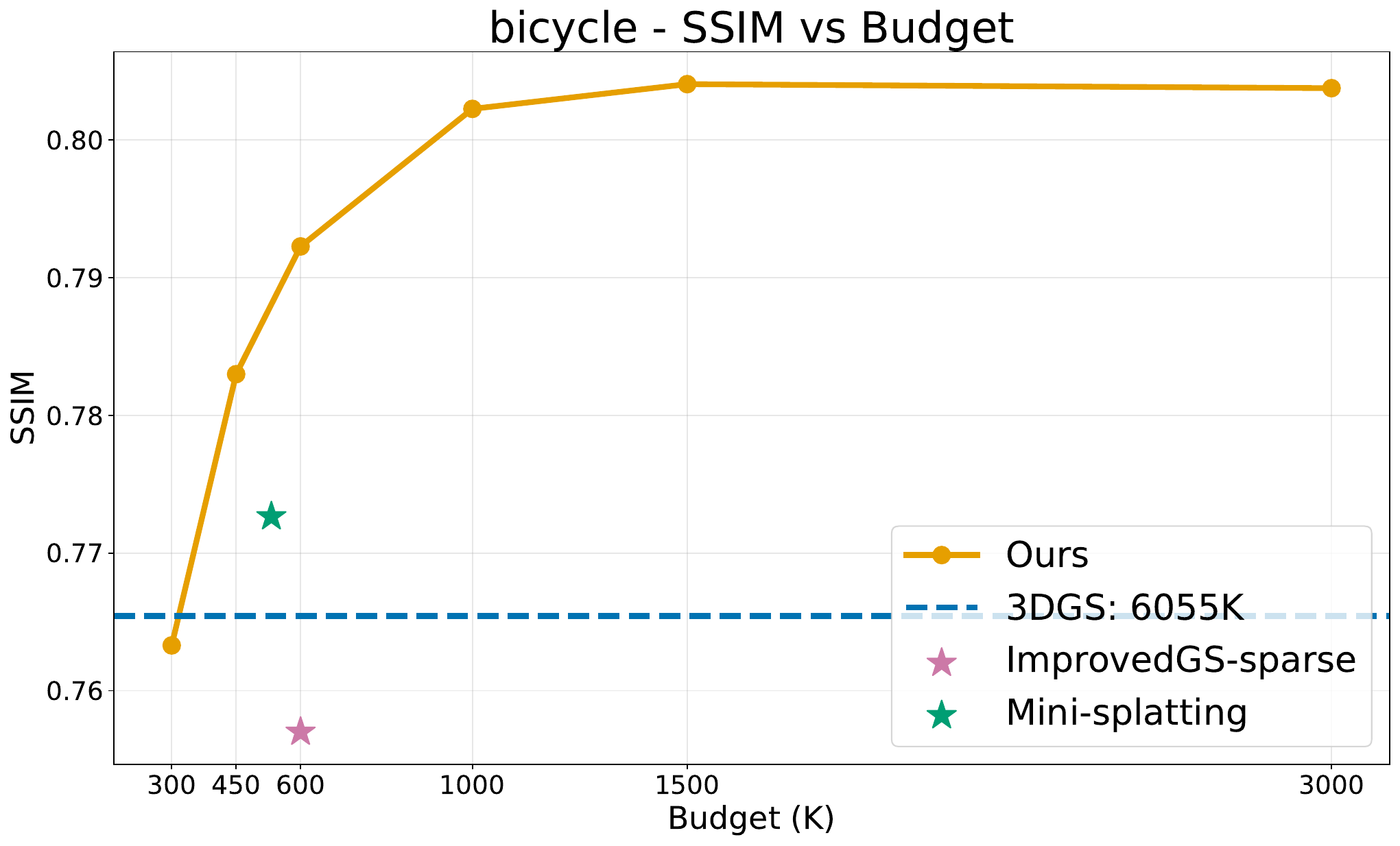}
\end{minipage}
\hfill
\begin{minipage}[b]{0.32\textwidth}
\centering
\includegraphics[width=\textwidth]{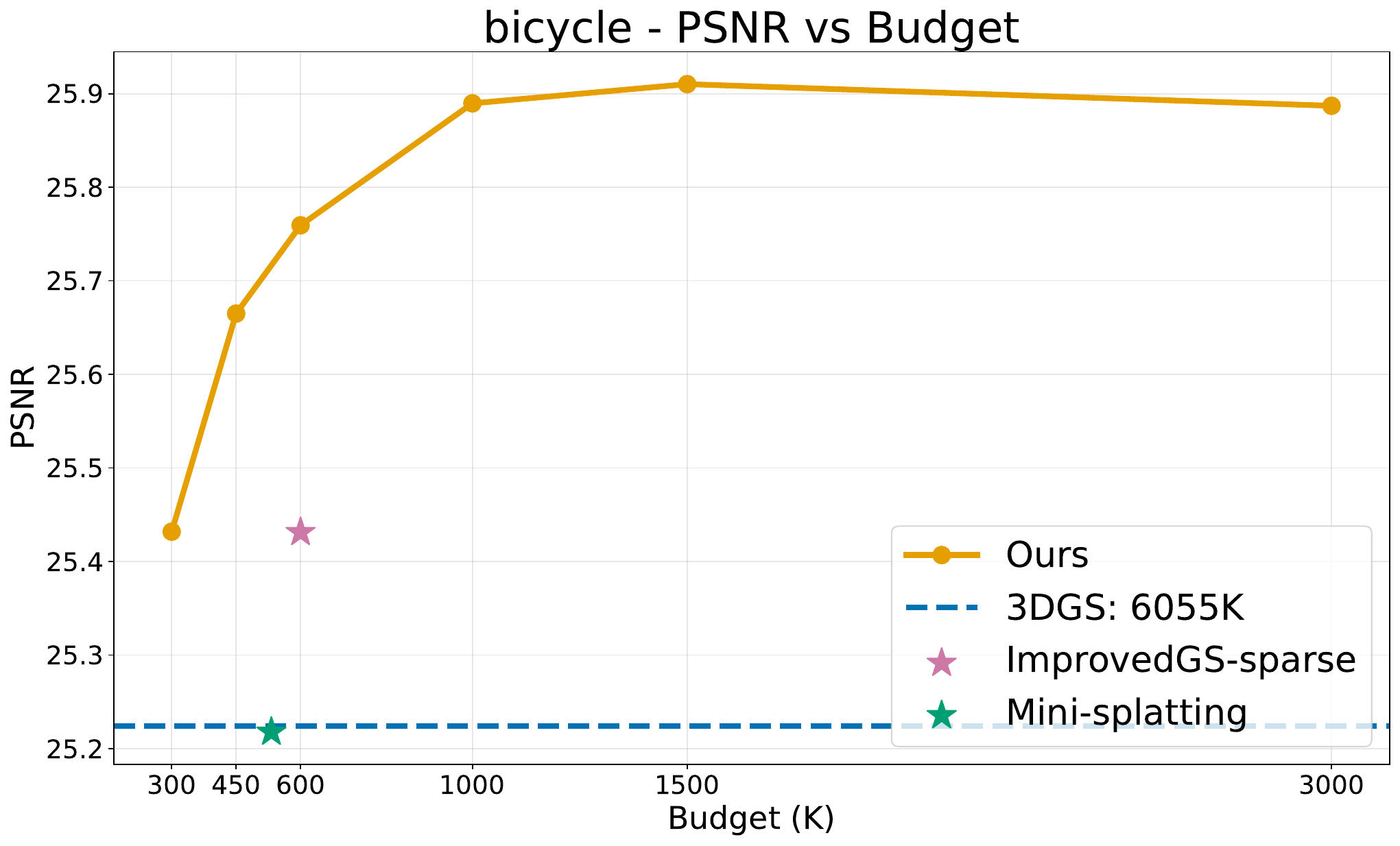}
\end{minipage}
\hfill
\begin{minipage}[b]{0.32\textwidth}
\centering
\includegraphics[width=\textwidth]{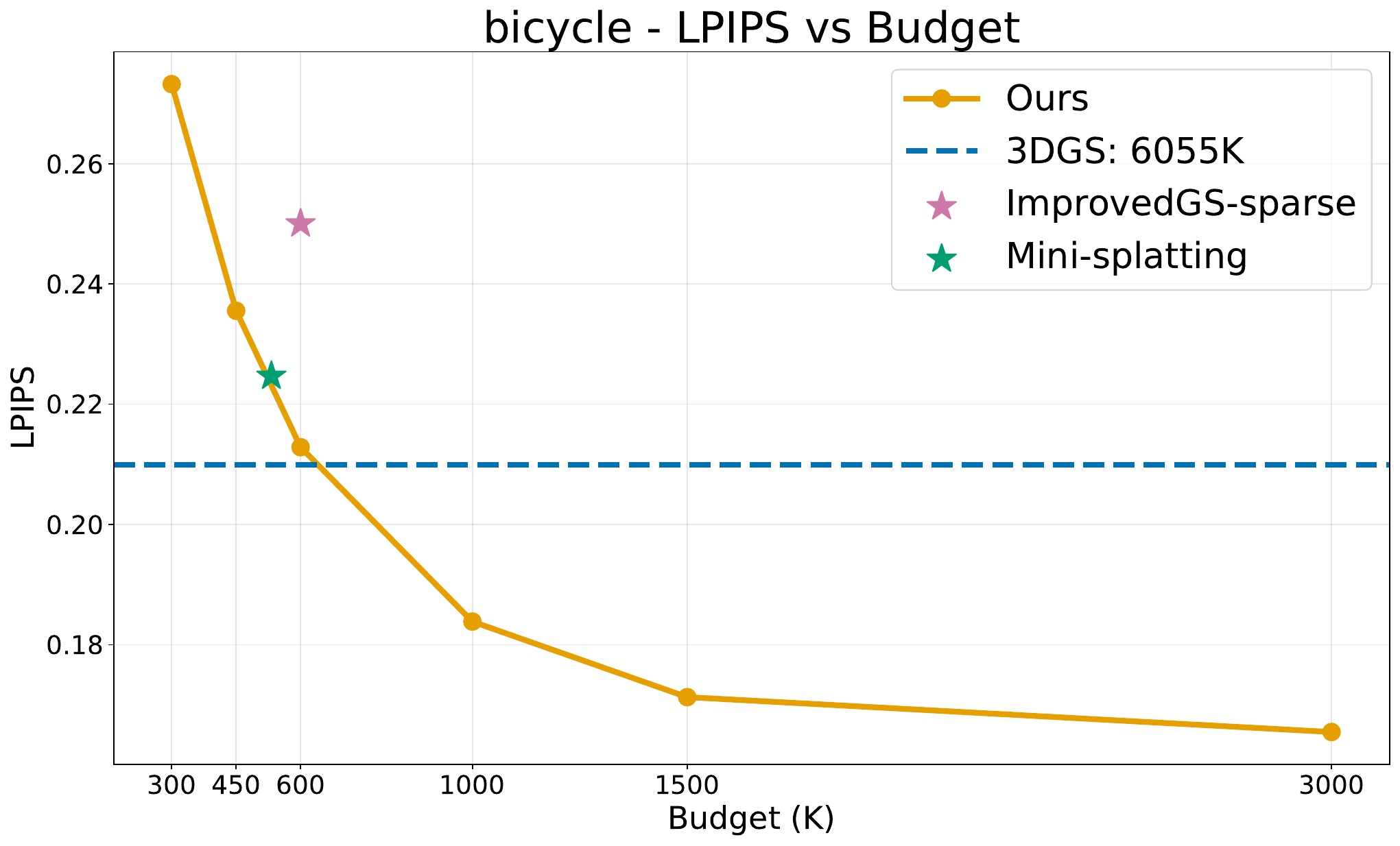}
\end{minipage}
\caption{Performance under varying Gaussian budgets. Results for the remaining scenes can be found in appendix.}
\label{fig:score-budget}
\end{figure*}

In the natural-selection framework, environmental pressure should be applied uniformly to all Gaussian opacities to guarantee fairness of the selection process. 
We define the opacity decay ratio: 
\begin{equation}
R_o = \frac{\alpha_{t-1}-\alpha_t}{\alpha_{t-1}}
\end{equation}
to quantify the degree of influence that the environmental pressure exerts on an individual. 
Ideally, applying the same decay ratio to every Gaussian accurately models the environmental pressure. 
Because the regularization gradient received by every Gaussian has the same magnitude and direction, it can be regarded as a gradient field. 
However, if, by the end of the natural-selection stage, the opacities of surviving Gaussians are generally suppressed too strongly, this will substantially degrade subsequent optimization convergence speed and numerical stability.

To satisfy Goal 2, we introduce a finite prior designed to accelerate the overall process without breaking the fairness of natural selection. 
There exists a correlation between a Gaussian’s current opacity and its fitness: high-opacity Gaussians tend to contribute more to rendering and are more likely to be important primitives; moreover, in the late stage of natural selection opacity already reflects fitness. 
We exploit this weak correlation as a limited prior to speed up both selection and the optimization recovery (see Figure~\ref{fig:prior}). 
Concretely, we introduce the finite prior by applying the gradient field to the pre-activation opacity parameter \(v\) instead of to \(alpha\) directly, 

Opacity $\alpha$ is activated by $v$ through the sigmoid function:
\begin{equation}
    \alpha = S(v) = \frac{1}{1 + e^{-v}}.
\end{equation}
Let the global regularization gradient be \( \nabla v \) and the learning rate be \( lr \). Ignoring rendering gradients, the update rule is:
\begin{equation}
    v_{t+1} = v_t + \nabla v \cdot lr,
\end{equation}
yielding the decay ratio:
\begin{equation}
    R_o^{t+1} = \frac{S(v_t) - S(v_t + \nabla v \cdot lr)}{S(v_t)}.
\end{equation}
When \( \nabla v \cdot lr \to 0 \),
\begin{equation}
    R_o^{t+1} \approx (1 - \alpha) \cdot |\nabla v \cdot lr|.
\end{equation}
Thus, the decay ratio \( R_o \) is linearly proportional to \( 1 - \alpha \):

\begin{itemize}
    \item As \( \alpha \to 1 \), \( R_o^{t+1} \to 0 \), protecting consistently high-fitness Gaussians;
    \item As \( \alpha \to 0 \), \( R_o^{t+1} \to |\nabla v \cdot lr| \), meaning the decay ratio approaches its maximum value. For example, compared to Gaussians with average opacity \( \alpha = 0.5 \) (whose decay ratio is \( 0.5 \cdot |\nabla v \cdot lr| \)), low-opacity Gaussians (\( \alpha \to 0 \)) exhibit up to twice the decay magnitude. This accelerates the elimination of less fit individuals.
\end{itemize}
This limited prior speeds up convergence while preserving the fairness and adaptivity of natural selection.

\begin{table*}[t]
	\centering
	\scalebox{0.74}{
		\begin{tabular}{l|ccccc|ccccc|ccccc}
			
			Dataset & \multicolumn{5}{c|}{Mip-NeRF360}  & \multicolumn{5}{c|}{Deep Blending} & \multicolumn{5}{c}{Tanks\&Temples} \\
			Method|Metric
			& $SSIM^\uparrow$   & $PSNR^\uparrow$    & $LPIPS^\downarrow$  & $Num^\downarrow$ & $Time^\downarrow$
			& $SSIM^\uparrow$   & $PSNR^\uparrow$    & $LPIPS^\downarrow$  & $Num^\downarrow$ & $Time^\downarrow$
			& $SSIM^\uparrow$   & $PSNR^\uparrow$    & $LPIPS^\downarrow$  & $Num^\downarrow$ & $Time^\downarrow$\\
			\hline
			3DGS & 0.816  & 27.50  & \colorbox{orange!40}{0.216}  & 3320453  & 35.9  & 0.904  & 29.56  & 0.244  & 2819180  & 32.5  & 0.849  & 23.72  & 0.177  & 1835092  & 19.9 \\
            \hline
            Compact-3DGS & 0.807  & 27.33  & 0.227  & 1516172  & 31.3  & 0.904  & 29.61  & 0.249  & 1251516  & 28.9  & 0.847  & 23.67  & 0.180  & 933650  & 18.0 \\
            Mini-splatting & \colorbox{orange!40}{0.822}  & 27.36  & 0.217  & 493466  & 26.8  & 0.910  & 30.07  & \colorbox{orange!40}{0.241}  & 554179  & 23.2  & 0.847  & 23.47  & 0.180  & \colorbox{red!40}{300481}  & 17.3 \\
            MaskGS & 0.815  & 27.43  & 0.218  & 1582926  & 31.9  & 0.907  & 29.76  & 0.245  & 910162  & 27.6  & 0.847  & 23.73  & 0.180  & 740683  & 17.4 \\
            GaussianSPA & 0.817  & 27.31  & 0.229  & \colorbox{red!40}{421427}  & 34.3  & \colorbox{orange!40}{0.913}  & 30.00  & 0.242  & \colorbox{red!40}{443740}  & 28.8  & 0.850  & 23.40  & \colorbox{orange!40}{0.171}  & \colorbox{orange!40}{424801}  & 23.3 \\
            ImprovedGS & 0.814  & \colorbox{orange!40}{27.66}  & 0.233  & \colorbox{orange!40}{466667}  & \colorbox{red!40}{8.1}  & 0.911  & \colorbox{red!40}{30.23}  & 0.244  & \colorbox{orange!40}{450000}  & \colorbox{red!40}{6.8}  & \colorbox{orange!40}{0.856}  & \colorbox{orange!40}{24.39}  & 0.179  & 450000  & \colorbox{red!40}{5.7} \\
            \hline
            Ours & \colorbox{red!40}{0.833}  & \colorbox{red!40}{28.13}  & \colorbox{red!40}{0.207}  & \colorbox{orange!40}{466667}  & \colorbox{orange!40}{12.5}  & \colorbox{red!40}{0.914}  & \colorbox{orange!40}{30.15}  & \colorbox{red!40}{0.233}  & \colorbox{orange!40}{450000}  & \colorbox{orange!40}{9.4}  & \colorbox{red!40}{0.871}  & \colorbox{red!40}{24.63}  & \colorbox{red!40}{0.154}  & 450000  & \colorbox{orange!40}{8.5} \\
        \end{tabular}
	}
	\caption{Quantitative results on the Mip-NeRF 360, Deep Blending, and Tanks and Temples datasets. Cells are highlighted as follows: \colorbox{red!40}{best}, and \colorbox{orange!40}{second best}. Per-scene detailed data can be found in the appendix.}
	\label{tab:quantitative_results}
\end{table*}

\textbf{Implementation Details:}
To prevent Gaussians with extremely high opacity from evading selection during the initial phase, we apply prior-free regularization under a low learning rate in the early stages of natural selection. After introducing the prior, a higher learning rate is subsequently used to accelerate the screening process.
The regularization gradient \( \nabla v \) is derived from:
\begin{equation}
    \mathcal{L}_{\text{reg}} = (\mathbb{E}[v] - T)^2,
\end{equation}
where \( \mathbb{E}[v] \) denotes the mean value of all $v$ parameters and \( T \) is the regularization target. Its gradient is:
\begin{equation}
    \nabla v = 2(\mathbb{E}[v] - T),
\end{equation}
When \( T \gg \mathbb{E}[v] \), \( \nabla v \) is primarily dominated by \( T \), remaining stable throughout the process. In practice, we set \( T = -20 \). Since the natural selection process terminates when the number of surviving Gaussians meets the budget requirement, \( T \) here primarily serves to provide a gradient direction and a stable magnitude. Because \( \nabla v \propto T \) (when \( |T| \gg |\mathbb{E}[v]| \)), increasing \( |T| \) is equivalent to linearly amplifying the strength of the gradient field, an effect similar to increasing the learning rate \( lr \).
\section{Experiments}

\subsection{Datasets and Metrics}

Following 3DGS and other compact 3DGS baselines, we select 13 scenes from the Mip-NeRF 360, Deep Blending, and Tanks and Temples datasets for evaluation. 
In each experiment, every 8th image is used as the test set, and the remaining images form the training set.

For quantitative evaluation, we adopt three widely used metrics: Peak Signal-to-Noise Ratio (PSNR), Structural Similarity Index (SSIM), and Learned Perceptual Image Patch Similarity (LPIPS). All metrics are computed using the official evaluation framework provided by 3DGS.

\subsection{Implementation Details}

Our method is independent of specific CUDA kernels and introduces no additional trainable parameters, making it easily transferable to advanced 3DGS variants. 
We adopt Improved-GS as the base model, since it provides faster and higher-quality densification. 
To reduce the influence of initial opacity before natural selection, we increase the opacity learning rate to $4$ times the original value, ensuring that survival is primarily determined by dynamic gradient competition.
The opacity learning rate will be restored to its original value after natural selection is completed.

The natural selection stage begins after densification (at 15K iterations) and continues until the number of remaining Gaussians meets the target budget. 
Empirically, this process converges within 5K–8K iterations, achieving the best rendering quality. Based on this observation, we tune the regularization learning rate per scene, while all other hyperparameters remain shared across scenes. 
\textbf{A detailed discussion of hyperparameter choices is provided in appendix.}

All experiments are conducted on a single RTX A5000 GPU, with a total of 30K training iterations for all methods. Competing methods are reimplemented using the official configurations provided by their authors. During the reproduction of GaussianSPA, several non-trivial issues were observed and are discussed in the appendix.
As noted in Section~\ref{sec:objective}, densification alone cannot meet the compact 3DGS requirement. 
Thus, we also include a sparse variant of Improved-GS—obtained by lowering the peak densification budget—as an additional baseline.

\subsection{Quantitative Comparison}
\label{sec:quantitative_omparison}

The quantitative results are summarized in Table~\ref{tab:quantitative_results}. Despite using a minimal Gaussian budget, our method achieves the best overall performance across PSNR, SSIM, and LPIPS. Compared with 3DGS, our approach attains a $>0.6 dB$ gain in PSNR while using only 15\% of the Gaussian budget. Relative to Improved-GS, our method shows comparable training time but significantly better rendering quality.

Mini-Splatting reaches a level similar to 3DGS with low budget usage, whereas GaussianSPA, due to its ADMM-based optimization, requires substantially longer iterations and performs pruning only at the end, resulting in longer training time. Compared to CompactGS, MaskGS achieves comparable or superior rendering quality with a similar or smaller budget.
Overall, our method achieves SOTA performance in compact 3DGS.

Figure~\ref{fig:score-budget} illustrates performance curves under different budget settings. 
Notably, in some scenes, our approach surpasses the original 3DGS using only 5\% of its budget. 

\subsection{Qualitative Comparison}

\begin{figure*}[t]
    \centering
    \includegraphics[width=0.9\textwidth]{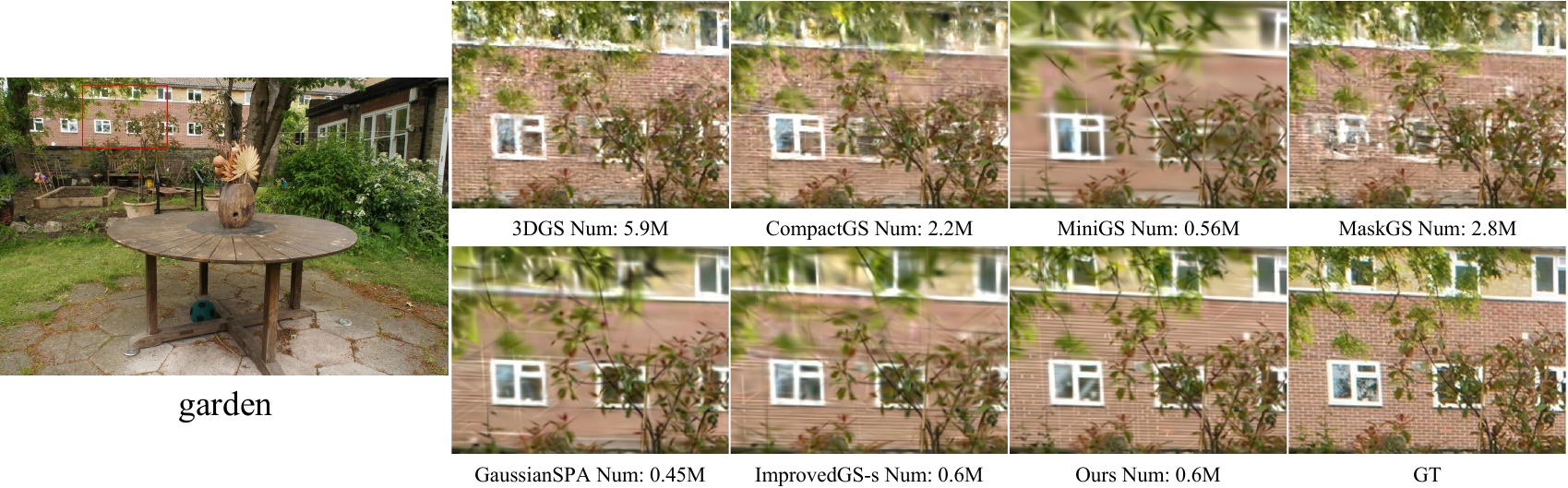} \\[10pt]
    \includegraphics[width=0.9\textwidth]{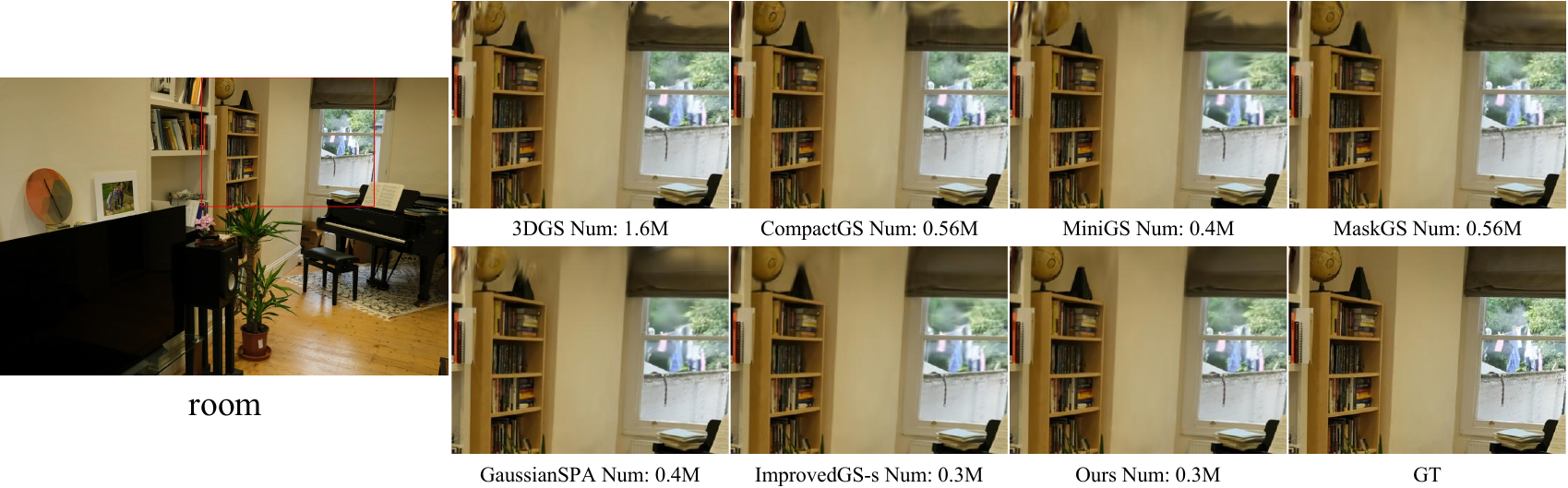} \\[10pt]
    \includegraphics[width=0.9\textwidth]{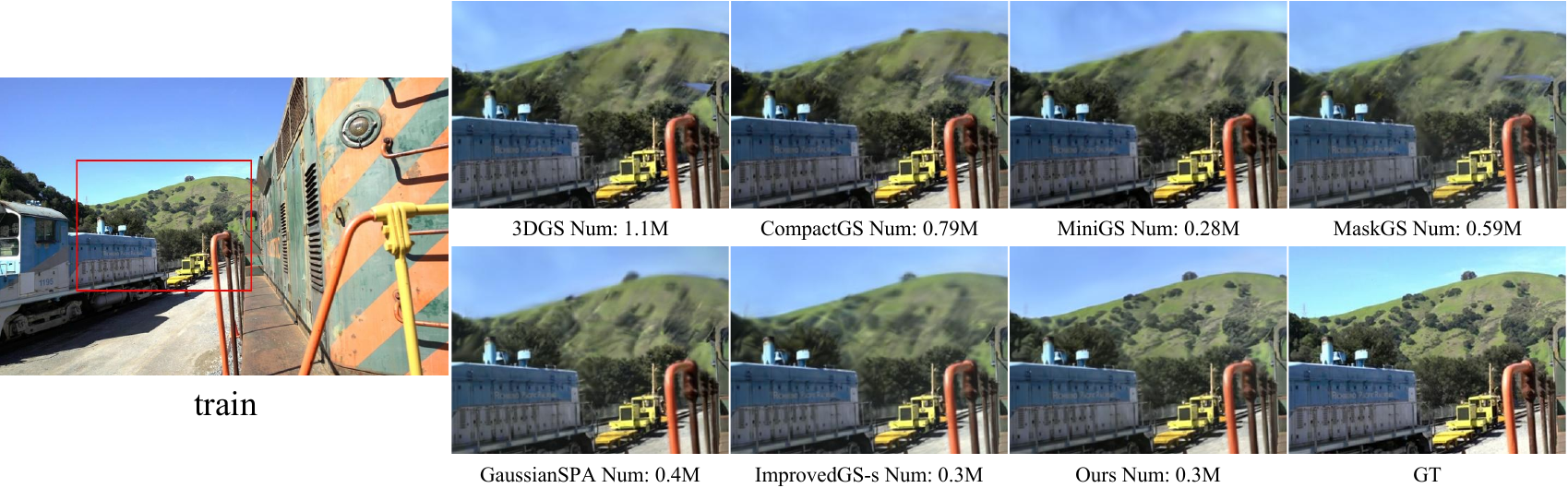}
    \caption{Qualitative comparison results among scenes garden, room, train. Num is final Gaussian count after training. The results for the remaining scenes can be found in appendix.}
    \label{fig:qualitative_comparison}
\end{figure*}

Representative qualitative results are shown in Figure~\ref{fig:qualitative_comparison}.

\begin{table}[t]
	\centering
	\scalebox{0.68}{
		\begin{tabular}{l|ccc|ccc}
			
			Dataset & \multicolumn{3}{c|}{Outdoors}  & \multicolumn{3}{c}{Indoors}\\
			Method|Metric
			& $SSIM^\uparrow$   & $PSNR^\uparrow$    & $LPIPS^\downarrow$
			& $SSIM^\uparrow$   & $PSNR^\uparrow$    & $LPIPS^\downarrow$\\
			\hline
			ImprovedGS (sparse) & 0.728  & 24.85  & 0.260  & 0.922  & 31.18  & 0.198 \\
            \hline
            SPA Pruning& 0.733  & 24.95  & 0.261  & \colorbox{orange!40}{0.928}  & \colorbox{orange!40}{31.70}  & \colorbox{orange!40}{0.184} \\
            MaskGS Pruning& \colorbox{orange!40}{0.749}  & \colorbox{orange!40}{25.11}  & \colorbox{orange!40}{0.242}  & 0.927  & 31.67  & \colorbox{orange!40}{0.184} \\
            \hline
            Opacity Pruning & 0.722  & 24.75  & 0.275  & 0.924  & 31.52  & 0.190 \\
            Render Pruning & 0.726  & 24.67  & 0.261  & 0.923  & 31.39  & 0.191 \\
            Edge Pruning & 0.728  & 24.81  & 0.263  & 0.924  & 31.50  & 0.193 \\
            \hline
            Natural Selection (Ours) & \colorbox{red!40}{0.753}  & \colorbox{red!40}{25.20}  & \colorbox{red!40}{0.234}  & \colorbox{red!40}{0.930}  & \colorbox{red!40}{31.78}  & \colorbox{red!40}{0.178} 
        \end{tabular}
	}
	\caption{Ablation Study of unified base model. }
	\label{tab:ablation_results}
\end{table}

\begin{table}[t]
	\centering
	\scalebox{0.7}{
		\begin{tabular}{l|ccc|ccc}
			Dataset & \multicolumn{3}{c|}{Outdoors}  & \multicolumn{3}{c}{Indoors}\\
			Method|Metric
			& $SSIM^\uparrow$   & $PSNR^\uparrow$    & $LPIPS^\downarrow$
			& $SSIM^\uparrow$   & $PSNR^\uparrow$    & $LPIPS^\downarrow$\\
			\hline
			No Prior & \colorbox{orange!40}{0.751} & \colorbox{orange!40}{25.18} & \colorbox{orange!40}{0.242} & 0.928 & 31.70 & 0.183 \\
            Strong Prior & 0.747 & 25.10 & 0.245 & \colorbox{orange!40}{0.929} & \colorbox{orange!40}{31.75} & \colorbox{orange!40}{0.181} \\
            \hline
            Finite Prior (Ours) & \colorbox{red!40}{0.753} & \colorbox{red!40}{25.20} & \colorbox{red!40}{0.234} & \colorbox{red!40}{0.930} & \colorbox{red!40}{31.78} & \colorbox{red!40}{0.178} 
        \end{tabular}
	}
	\caption{Ablation Study of prior. }
	\label{tab:ablation_prior}
\end{table}

\textbf{Garden:}
Our method achieves higher rendering quality than 3DGS while using only 10\% of the Gaussian budget. Although Mini-Splatting and GaussianSPA consume less budget, their reconstruction quality is far inferior. Compared to ImprovedGS, our approach restores finer scene details and avoids large blurry regions.

\textbf{Room:}
mong all methods, our approach produces fewer artifacts and is the only one that successfully restores the details of the globe in the upper left corner.
\textbf{Train:}
Our method delivers superior reconstruction quality while maintaining the lowest budget among all methods. 
Except for ImprovedGS-sparse, all other baselines exhibit visible artifacts and fail to reconstruct the distant mountain contours accurately. Our approach produces more faithful geometry and textures.

\subsection{Ablation Studies}

\begin{figure*}[ht]
    \centering
    \includegraphics[width=0.92\textwidth]{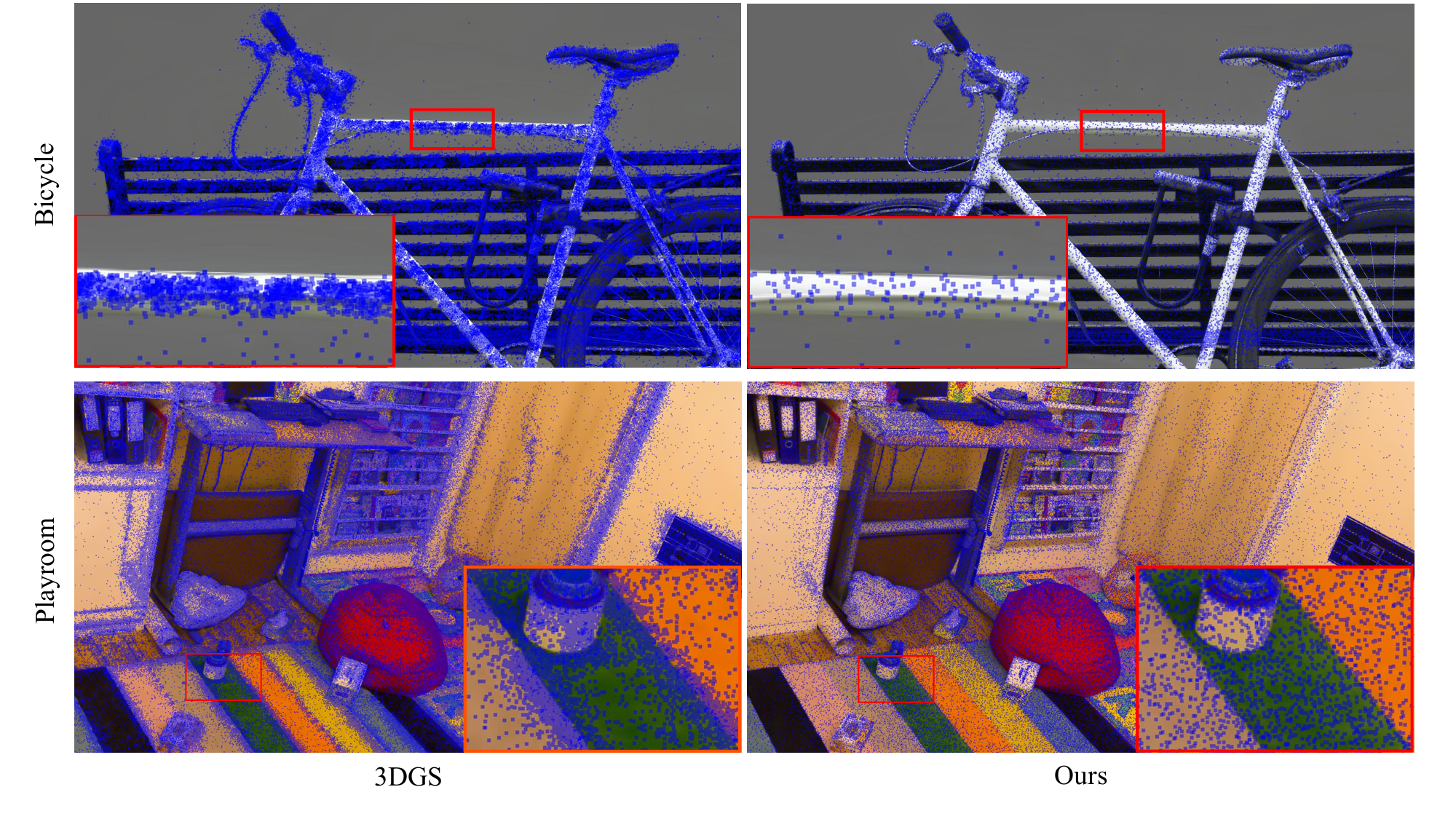}
    \caption{Comparison of the point cloud distributions in scenes trained with 3DGS and our method. For the bicycle scene, the background has been removed to more clearly highlight the differences between the two approaches.}
    \label{fig:points}
\end{figure*}

\textbf{Unified Base Model:}
To eliminate the influence of inherent model differences (e.g., Improved-GS, 3DGS, MiniGS-D) and ensure a fair comparison of pruning techniques, we apply all pruning methods on a unified Improved-GS (15K iteration) base model.
We choose MaskGS as the representative of mask-based pruning and SPA for smooth pruning. Additionally, we test three non-learnable pruning weights: (1) opacity-based weight (used in LightGaussian and Mini-Splatting), (2) rendering-weight, and (3) edge-weight from Improved-GS densification. These weights are used as relative retention probabilities to prevent structural voids.

As shown in Table~\ref{tab:ablation_results}, our method consistently outperforms all others. Among the baselines, MaskGS performs better in outdoor scenes, while SPA is slightly superior in indoor scenes. This indicates that in complex scenes, selecting the right Gaussians is critical, whereas in simpler scenes, smoothness of pruning plays a larger role. Our approach effectively balances both aspects, achieving the best results.
We also observe that single-weight heuristics differ little among themselves but remain substantially worse than learnable methods.

\textbf{Finite Prior Ablation:}
To validate the effectiveness of the finite prior design, we conduct two ablation variants:

\begin{enumerate}
    \item No Prior: By manually compensating gradients, the opacity attenuation magnitude of all Gaussians is maintained strictly consistent, thereby eliminating prior influence. Since direct parameter modification invalidates the optimizer, manual gradient compensation serves as an approximation.
    \item Strong Prior: Using opacity as the sampling probability for exemption from natural selection in the current round.
\end{enumerate}

As shown in Table~\ref{tab:ablation_prior}, the No Prior variant fails to achieve efficient convergence (Goal 2) and performs poorly in indoor scenes. The Strong Prior variant disrupts the fairness and adaptability of natural selection (Goal 1), leading to significant performance drops in outdoor scenes.
Our finite prior strikes the optimal balance between fairness and efficiency. 
More ablation study results can be found in appendix.

\subsection{Spatial Distribution of Point Cloud}

Figure~\ref{fig:points} compares the spatial distribution of Gaussians between 3DGS and our method after training. 
Apart from a significant reduction in the total number of Gaussians, our method yields more uniform distributions. 
In 3DGS, Gaussian points often form dense clusters, as observed in the \textit{bicycle} scene, where many Gaussians are concentrated in low-frequency areas such as the bicycle frame, resulting in redundancy.
Mini-Splatting mitigates this problem via depth reinitialization, whereas our method eliminates redundancy without any external intervention. 
During training, the positions of Gaussians in edge regions undergo repeated fine-tuning, causing these areas to receive high gradients and form excessively dense distributions (e.g., along table edges and carpet boundaries in the \textit{Playroom} scene). 
Our method avoids such over-clustering effectively.

\section{Conclusion}
\label{sec:conclusion}
We presented a biologically inspired simplification framework for 3D Gaussian Splatting that leverages natural selection principles to autonomously prune redundant Gaussians.
By applying uniform survival pressure through opacity regularization and using optimization gradients as fitness indicators, our method eliminates the need for manual criteria or additional parameters. 
Accelerated by finite-prior opacity decay, it achieves SOTA rendering quality with only 15\% of the original Gaussian budget and one-third of the training time, while improving PSNR by over 0.6 dB. 
This efficient, portable solution significantly lowers computational barriers, enabling broader adoption of 3DGS in resource-constrained applications.
Future work will explore integrating our natural-selection framework into dynamic or sparse-view reconstruction.

\clearpage

{
    \small
    \bibliographystyle{ieeenat_fullname}
    \bibliography{main}
}

\newpage
\appendix
\onecolumn
\section{Appendix}
\setlength{\tabcolsep}{6pt}

\subsection{Quantitative Comparison per Scene}

\begin{table}[ht]
	\centering
    \scalebox{0.9}{
	\begin{tabular}{l|c|c|c|c|c|c|c}
		Scene|Methods & 3DGS & CompactGS & MiniGS & MaskGS & SPA & ImprovedGS-s & Ours \\
        \hline
        bicycle & 0.765  & 0.745  & 0.773  & 0.765  & 0.761  & 0.757  & 0.792 \\
        flowers & 0.606  & 0.592  & 0.625  & 0.604  & 0.609  & 0.610  & 0.640 \\
        garden & 0.867  & 0.855  & 0.848  & 0.866  & 0.840  & 0.838  & 0.867 \\
        stump & 0.773  & 0.756  & 0.805  & 0.774  & 0.796  & 0.790  & 0.813 \\
        treehill & 0.632  & 0.628  & 0.654  & 0.634  & 0.656  & 0.648  & 0.664 \\
        bonsai & 0.942  & 0.941  & 0.939  & 0.941  & 0.940  & 0.939  & 0.947 \\
        counter & 0.908  & 0.907  & 0.905  & 0.907  & 0.906  & 0.908  & 0.916 \\
        kitchen & 0.928  & 0.926  & 0.926  & 0.926  & 0.923  & 0.919  & 0.931 \\
        room & 0.919  & 0.918  & 0.921  & 0.919  & 0.922  & 0.921  & 0.929 \\
        \hline
        playroom & 0.907  & 0.908  & 0.914  & 0.910  & 0.916  & 0.913  & 0.916 \\
        drjohnson & 0.901  & 0.901  & 0.907  & 0.904  & 0.910  & 0.909  & 0.911 \\
        \hline
        train & 0.815  & 0.813  & 0.812  & 0.812  & 0.813  & 0.819  & 0.843 \\
        truck & 0.882  & 0.880  & 0.883  & 0.881  & 0.887  & 0.892  & 0.899 

        \end{tabular}}
	\caption{The SSIM scores for all works in each scene.}
\end{table}

\begin{table}[ht]
	\centering
    \scalebox{0.9}{
	\begin{tabular}{l|c|c|c|c|c|c|c}
		Scene|Methods & 3DGS & CompactGS & MiniGS & MaskGS & SPA & ImprovedGS-s & Ours \\
        \hline
        bicycle & 25.22  & 24.92  & 25.22  & 25.21  & 25.07  & 25.43  & 25.76 \\
        flowers & 21.62  & 21.39  & 21.53  & 21.55  & 21.56  & 21.59  & 21.86 \\
        garden & 27.41  & 27.13  & 26.91  & 27.39  & 26.80  & 27.17  & 27.79 \\
        stump & 26.64  & 26.31  & 27.24  & 26.66  & 27.05  & 27.05  & 27.38 \\
        treehill & 22.44  & 22.48  & 22.73  & 22.52  & 23.03  & 23.00  & 23.16 \\
        bonsai & 32.26  & 32.14  & 31.49  & 32.08  & 31.40  & 32.22  & 32.77 \\
        counter & 29.03  & 28.98  & 28.65  & 28.96  & 28.56  & 29.31  & 29.68 \\
        kitchen & 31.48  & 31.15  & 31.26  & 31.14  & 30.99  & 31.06  & 32.25 \\
        room & 31.43  & 31.49  & 31.25  & 31.40  & 31.36  & 32.14  & 32.52 \\
        \hline
        playroom & 29.96  & 30.06  & 30.57  & 30.15  & 30.49  & 30.66  & 30.63 \\
        drjohnson & 29.16  & 29.17  & 29.57  & 29.37  & 29.51  & 29.79  & 29.66 \\
        \hline
        train & 22.00  & 22.01  & 21.61  & 22.09  & 21.34  & 22.43  & 22.74 \\
        truck & 25.44  & 25.33  & 25.33  & 25.36  & 25.47  & 26.35  & 26.52
        \end{tabular}}
	\caption{The PSNR scores for all works in each scene.}
\end{table}

\begin{table}[ht]
	\centering
    \scalebox{0.9}{
	\begin{tabular}{l|c|c|c|c|c|c|c}
		Scene|Methods & 3DGS & CompactGS & MiniGS & MaskGS & SPA & ImprovedGS-s & Ours \\
        \hline
        bicycle & 0.210  & 0.235  & 0.225  & 0.212  & 0.251  & 0.250  & 0.213 \\
        flowers & 0.335  & 0.351  & 0.327  & 0.337  & 0.346  & 0.337  & 0.307 \\
        garden & 0.106  & 0.124  & 0.150  & 0.108  & 0.168  & 0.164  & 0.127 \\
        stump & 0.214  & 0.237  & 0.199  & 0.216  & 0.225  & 0.217  & 0.197 \\
        treehill & 0.327  & 0.335  & 0.313  & 0.328  & 0.331  & 0.333  & 0.306 \\
        bonsai & 0.203  & 0.205  & 0.200  & 0.206  & 0.199  & 0.215  & 0.193 \\
        counter & 0.200  & 0.204  & 0.198  & 0.203  & 0.198  & 0.206  & 0.189 \\
        kitchen & 0.126  & 0.128  & 0.129  & 0.129  & 0.134  & 0.150  & 0.126 \\
        room & 0.218  & 0.223  & 0.212  & 0.222  & 0.208  & 0.221  & 0.202 \\
        \hline
        playroom & 0.243  & 0.248  & 0.238  & 0.247  & 0.239  & 0.248  & 0.235 \\
        drjohnson & 0.244  & 0.249  & 0.243  & 0.243  & 0.245  & 0.240  & 0.230 \\
        \hline
        train & 0.207  & 0.210  & 0.222  & 0.212  & 0.216  & 0.219  & 0.190 \\
        truck & 0.146  & 0.151  & 0.139  & 0.149  & 0.126  & 0.139  & 0.119 \\

        \end{tabular}}
	\caption{The LPIPS scores for all works in each scene.}
\end{table}

\begin{table}[ht]
	\centering
    \scalebox{0.72}{
	\begin{tabular}{l|c|c|c|c|c|c|c}
		Scene|Methods & ImprovedGS-s & SPA	& MaskGS & Opacity Pruning	& Render Pruning & Edge Pruning	& Natural Selection(Ours)\\
        \hline
        bicycle & 0.757  & 0.759  & 0.781  & 0.751  & 0.755  & 0.757  & 0.792 \\
        flowers & 0.610  & 0.618  & 0.636  & 0.606  & 0.610  & 0.613  & 0.637 \\
        garden & 0.838  & 0.844  & 0.861  & 0.842  & 0.845  & 0.845  & 0.867 \\
        stump & 0.790  & 0.799  & 0.808  & 0.783  & 0.783  & 0.788  & 0.813 \\
        treehill & 0.648  & 0.647  & 0.658  & 0.628  & 0.637  & 0.636  & 0.657 \\
        bonsai & 0.939  & 0.945  & 0.944  & 0.941  & 0.939  & 0.941  & 0.947 \\
        counter & 0.908  & 0.913  & 0.912  & 0.908  & 0.907  & 0.908  & 0.916 \\
        kitchen & 0.919  & 0.926  & 0.928  & 0.925  & 0.925  & 0.924  & 0.930 \\
        room & 0.921  & 0.926  & 0.924  & 0.923  & 0.921  & 0.922  & 0.929 \\
        \end{tabular}}
	\caption{The SSIM scores for all ablation studies in each scene.}
\end{table}

\begin{table}[ht]
	\centering
    \scalebox{0.72}{
	\begin{tabular}{l|c|c|c|c|c|c|c}
		Scene|Methods & ImprovedGS-s & SPA	& MaskGS & Opacity Pruning	& Render Pruning & Edge Pruning	& Natural Selection(Ours)\\
        \hline
        bicycle & 25.43  & 25.38  & 25.63  & 25.25  & 25.25  & 25.39  & 25.80 \\
        flowers & 21.59  & 21.74  & 21.85  & 21.46  & 21.41  & 21.55  & 21.85 \\
        garden & 27.17  & 27.33  & 27.68  & 27.29  & 27.23  & 27.39  & 27.80 \\
        stump & 27.05  & 27.19  & 27.26  & 26.90  & 26.74  & 26.95  & 27.42 \\
        treehill & 23.00  & 23.12  & 23.15  & 22.87  & 22.71  & 22.78  & 23.16 \\
        bonsai & 32.22  & 32.74  & 32.63  & 32.40  & 32.19  & 32.46  & 32.74 \\
        counter & 29.31  & 29.61  & 29.59  & 29.42  & 29.47  & 29.55  & 29.69 \\
        kitchen & 31.06  & 32.08  & 32.08  & 31.96  & 31.81  & 31.74  & 32.22 \\
        room & 32.14  & 32.36  & 32.40  & 32.30  & 32.10  & 32.25  & 32.45 \\

        \end{tabular}}
	\caption{The PSNR scores for all ablation studies in each scene.}
\end{table}

\begin{table}[ht]
	\centering
    \scalebox{0.72}{
	\begin{tabular}{l|c|c|c|c|c|c|c}
		Scene|Methods & ImprovedGS-s & SPA	& MaskGS & Opacity Pruning	& Render Pruning & Edge Pruning	& Natural Selection(Ours)\\
        \hline
        bicycle & 0.250  & 0.260  & 0.236  & 0.265  & 0.253  & 0.256  & 0.214 \\
        flowers & 0.337  & 0.330  & 0.311  & 0.339  & 0.332  & 0.330  & 0.311 \\
        garden & 0.164  & 0.161  & 0.138  & 0.171  & 0.155  & 0.161  & 0.128 \\
        stump & 0.217  & 0.217  & 0.206  & 0.236  & 0.229  & 0.225  & 0.197 \\
        treehill & 0.333  & 0.338  & 0.320  & 0.364  & 0.336  & 0.344  & 0.318 \\
        bonsai & 0.215  & 0.196  & 0.198  & 0.205  & 0.207  & 0.206  & 0.194 \\
        counter & 0.206  & 0.194  & 0.196  & 0.204  & 0.203  & 0.205  & 0.189 \\
        kitchen & 0.150  & 0.135  & 0.130  & 0.137  & 0.135  & 0.141  & 0.126 \\
        room & 0.221  & 0.208  & 0.214  & 0.216  & 0.217  & 0.218  & 0.203 \\
        \end{tabular}}
	\caption{The LPIPS scores for all ablation studies in each scene.}
\end{table}

\subsection{Performance Under Varying Gaussian Budgets}

\begin{figure*}[t]
\centering
\begin{minipage}[b]{0.3\textwidth}
\centering
\includegraphics[width=\textwidth]{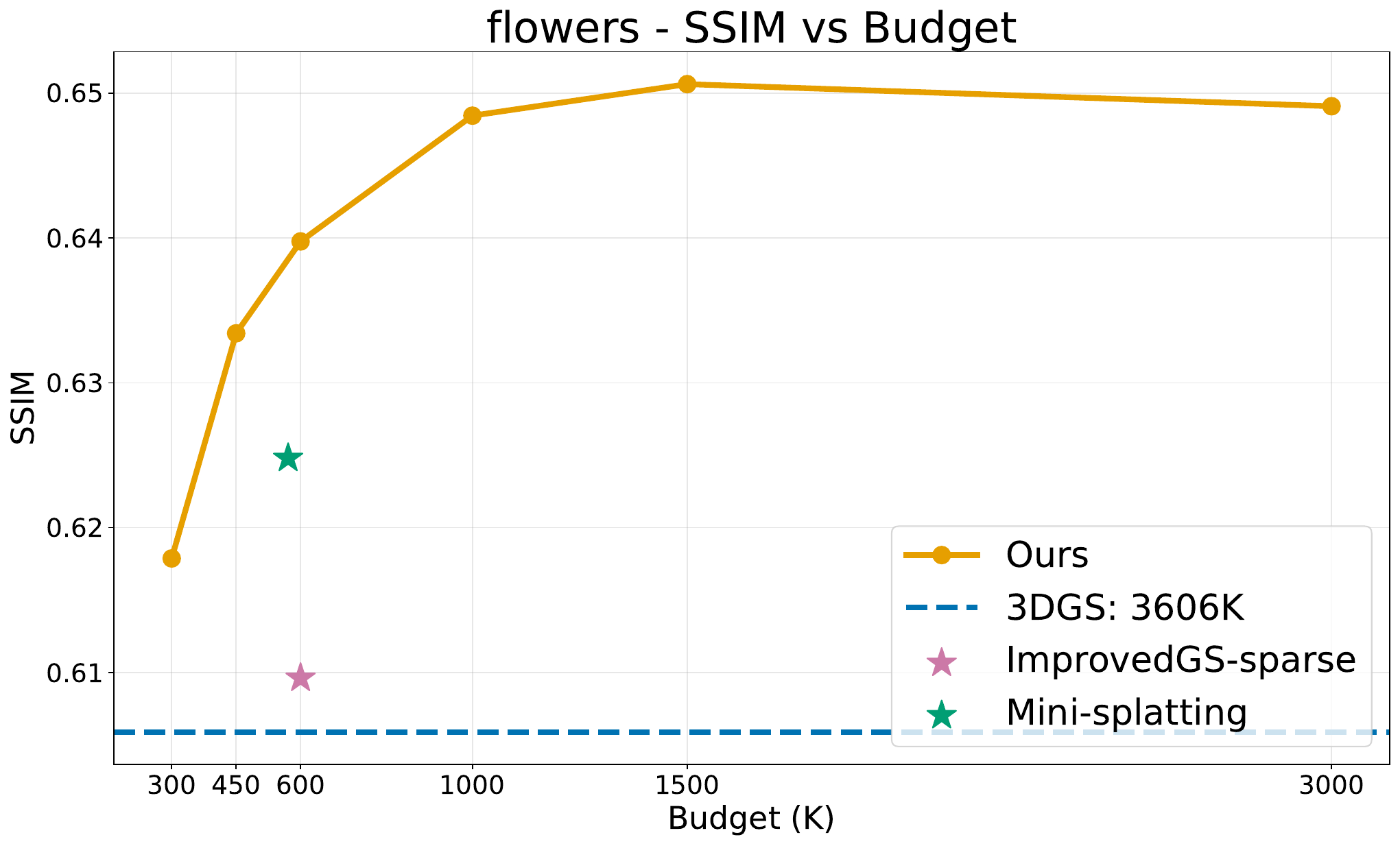}
\end{minipage}
\hfill
\begin{minipage}[b]{0.3\textwidth}
\centering
\includegraphics[width=\textwidth]{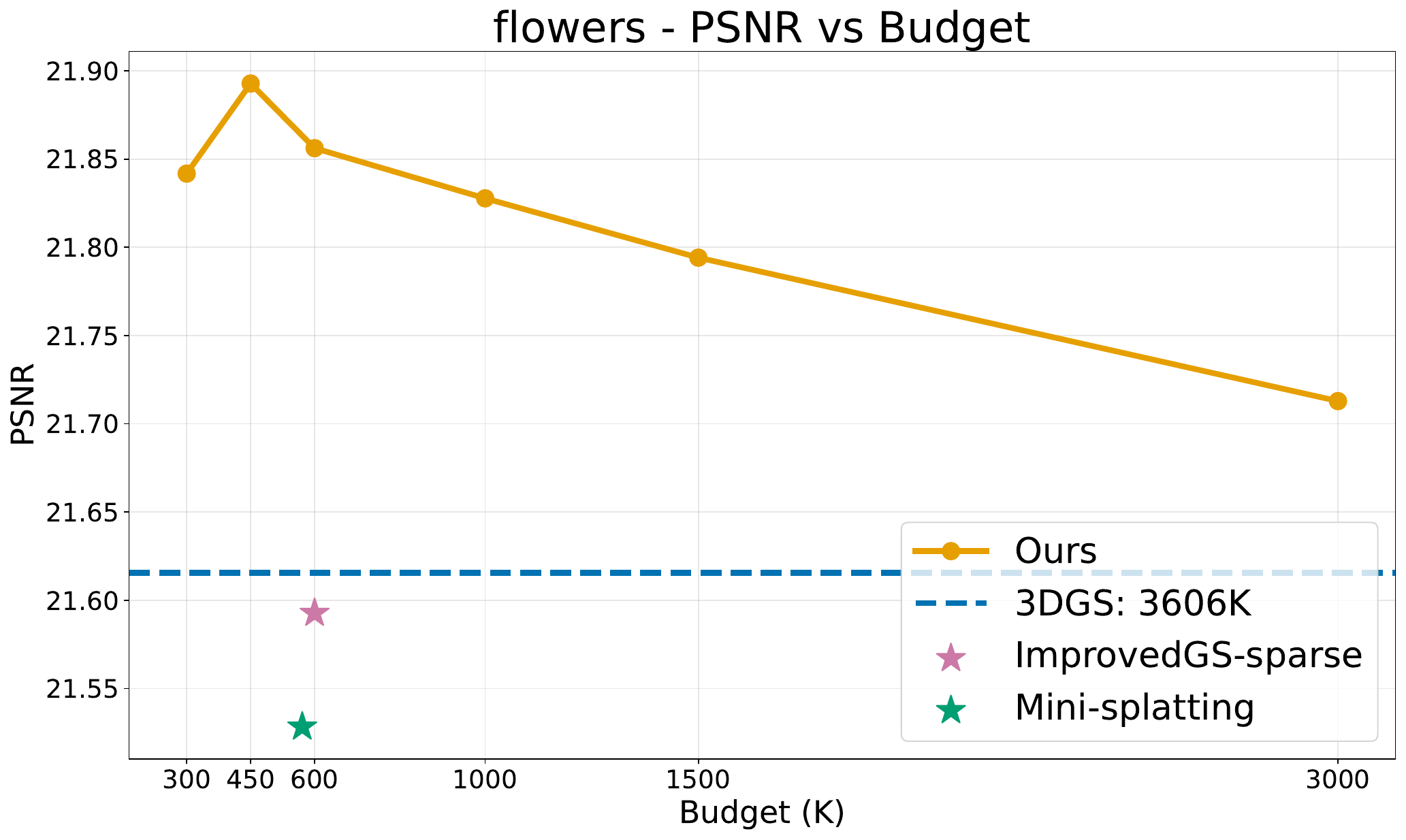}
\end{minipage}
\hfill
\begin{minipage}[b]{0.3\textwidth}
\centering
\includegraphics[width=\textwidth]{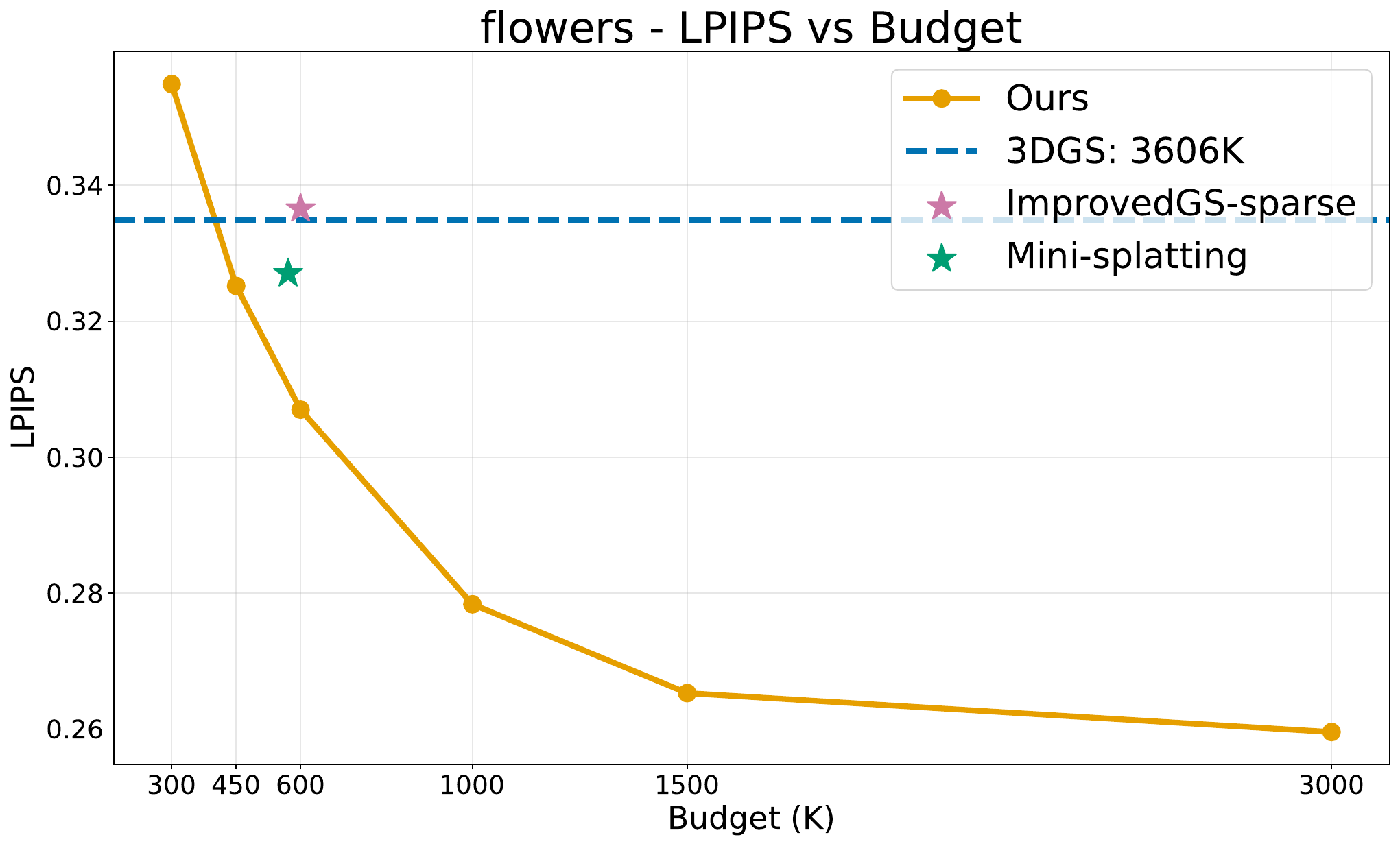}
\end{minipage}

\centering
\begin{minipage}[b]{0.3\textwidth}
\centering
\includegraphics[width=\textwidth]{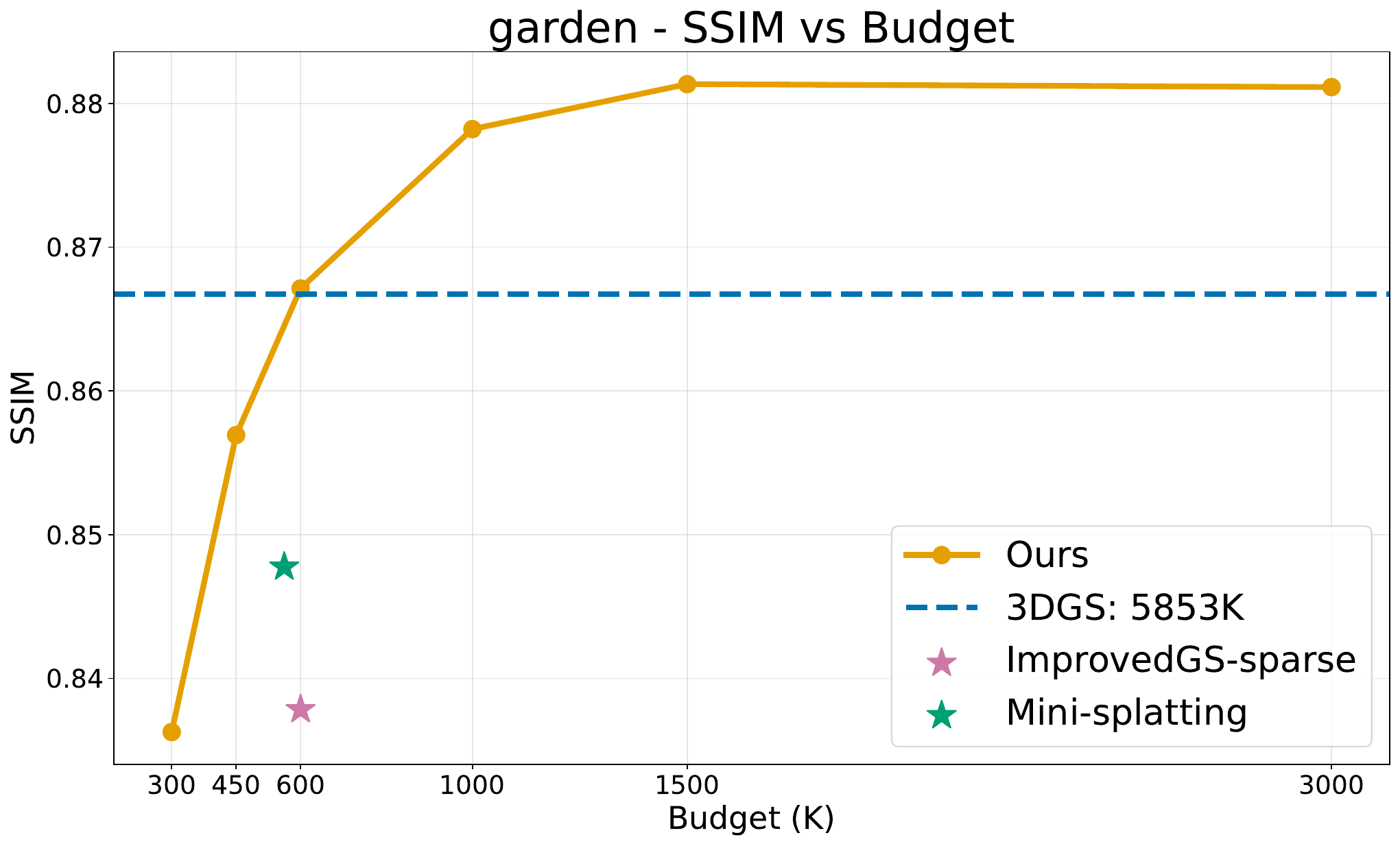}
\end{minipage}
\hfill
\begin{minipage}[b]{0.3\textwidth}
\centering
\includegraphics[width=\textwidth]{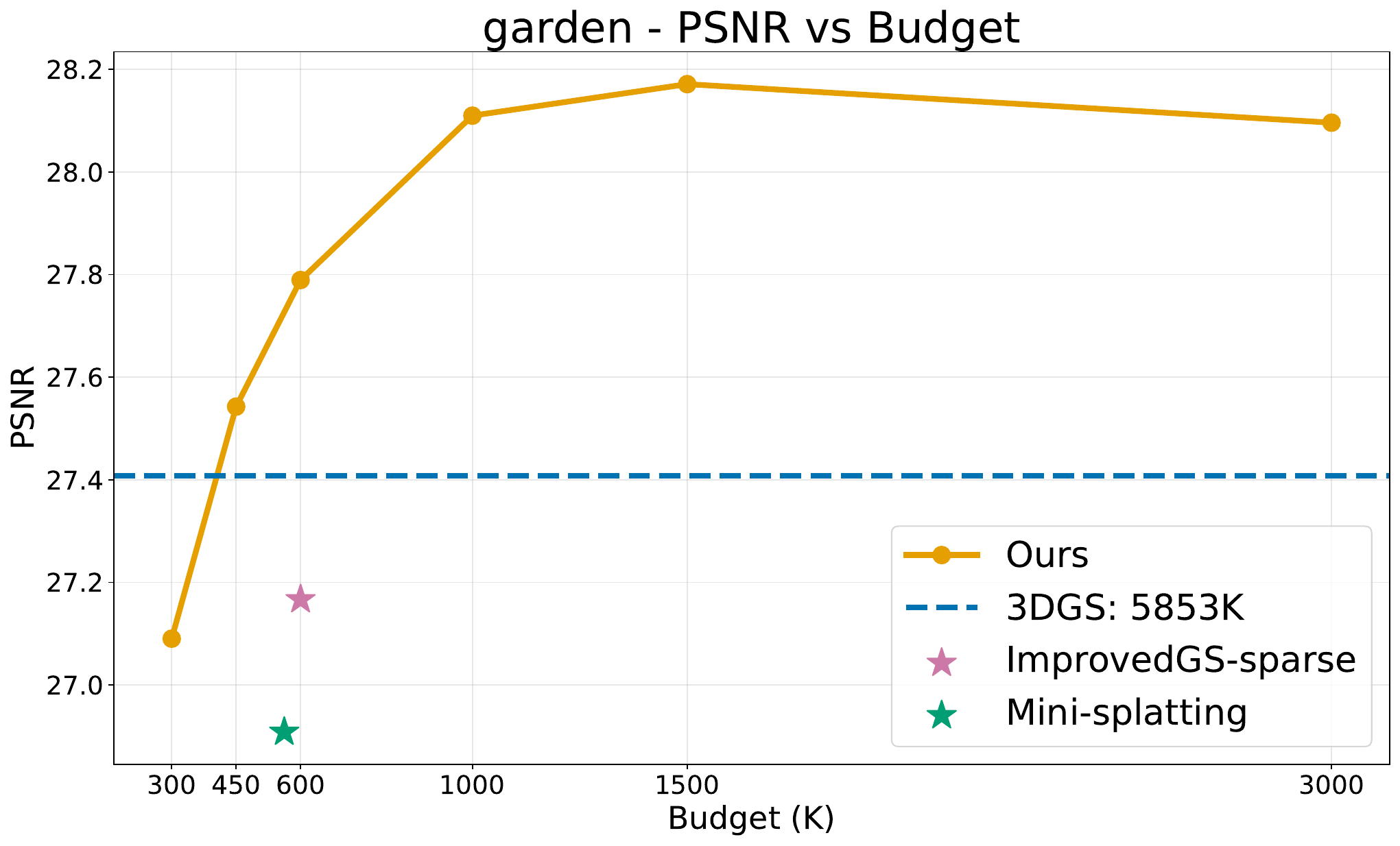}
\end{minipage}
\hfill
\begin{minipage}[b]{0.3\textwidth}
\centering
\includegraphics[width=\textwidth]{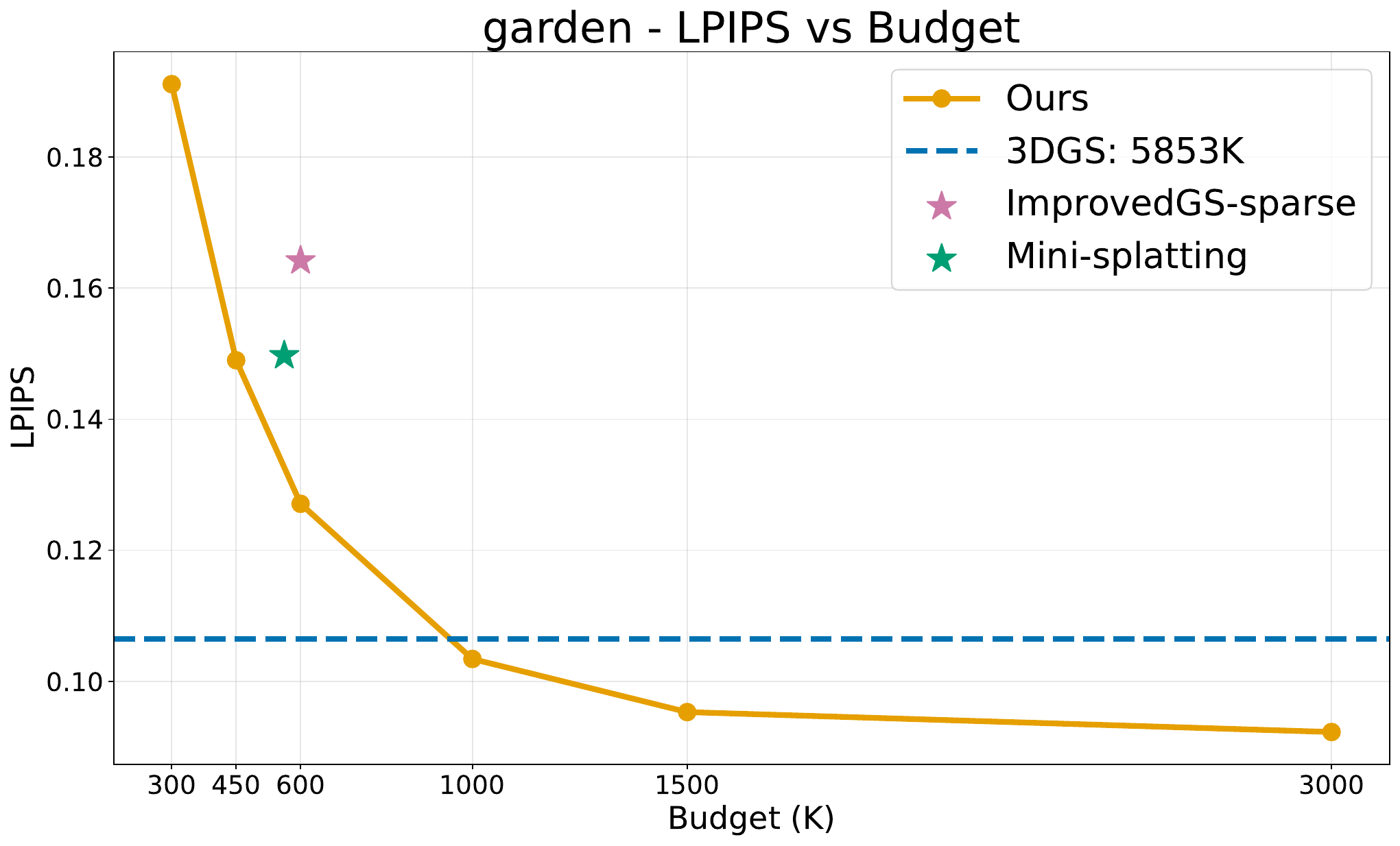}
\end{minipage}

\caption{Performance under varying Gaussian budgets of 2 scenes.}
\label{fig:score-budget2}
\end{figure*}

\begin{figure*}[t]
\centering
\begin{minipage}[b]{0.3\textwidth}
\centering
\includegraphics[width=\textwidth]{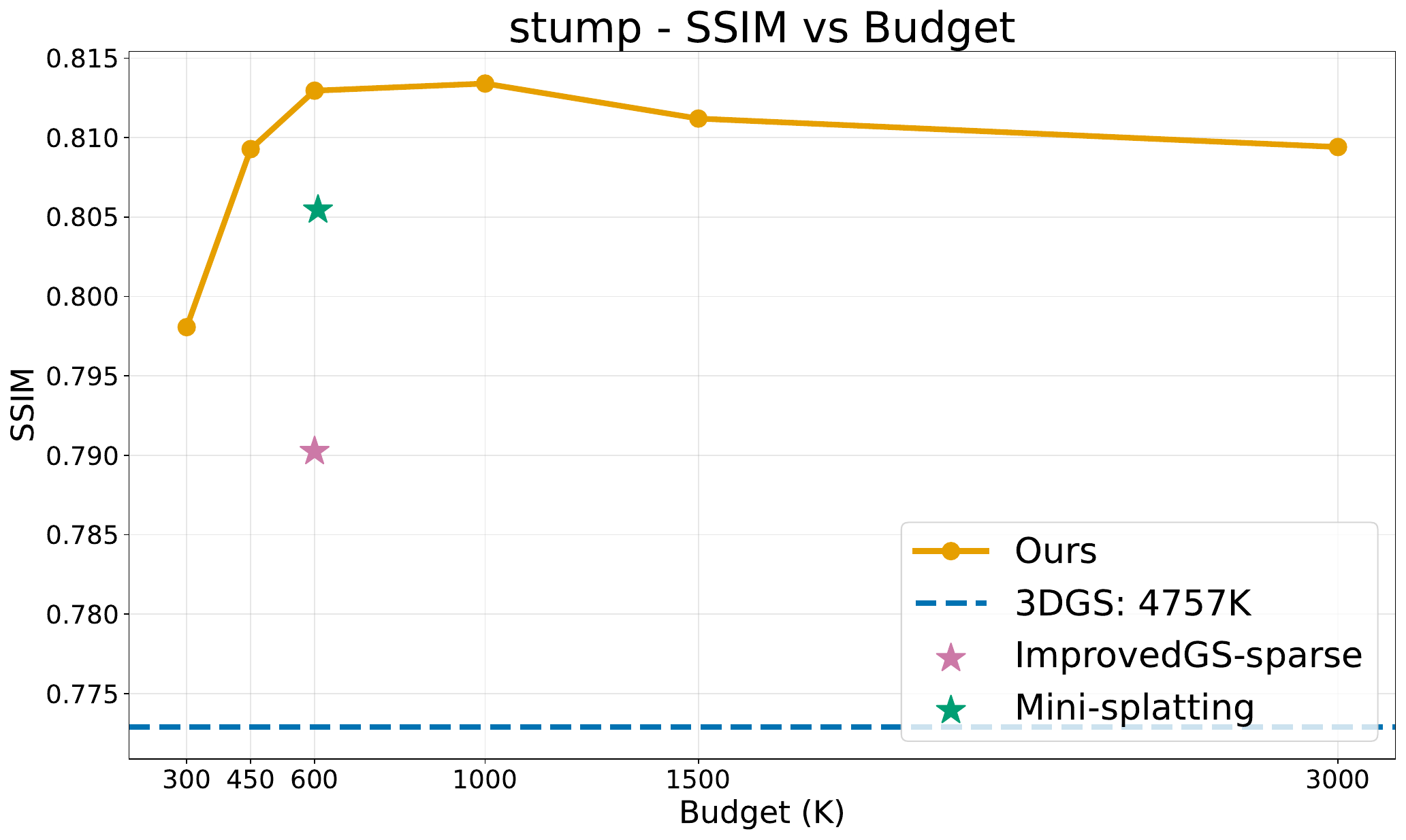}
\end{minipage}
\hfill
\begin{minipage}[b]{0.3\textwidth}
\centering
\includegraphics[width=\textwidth]{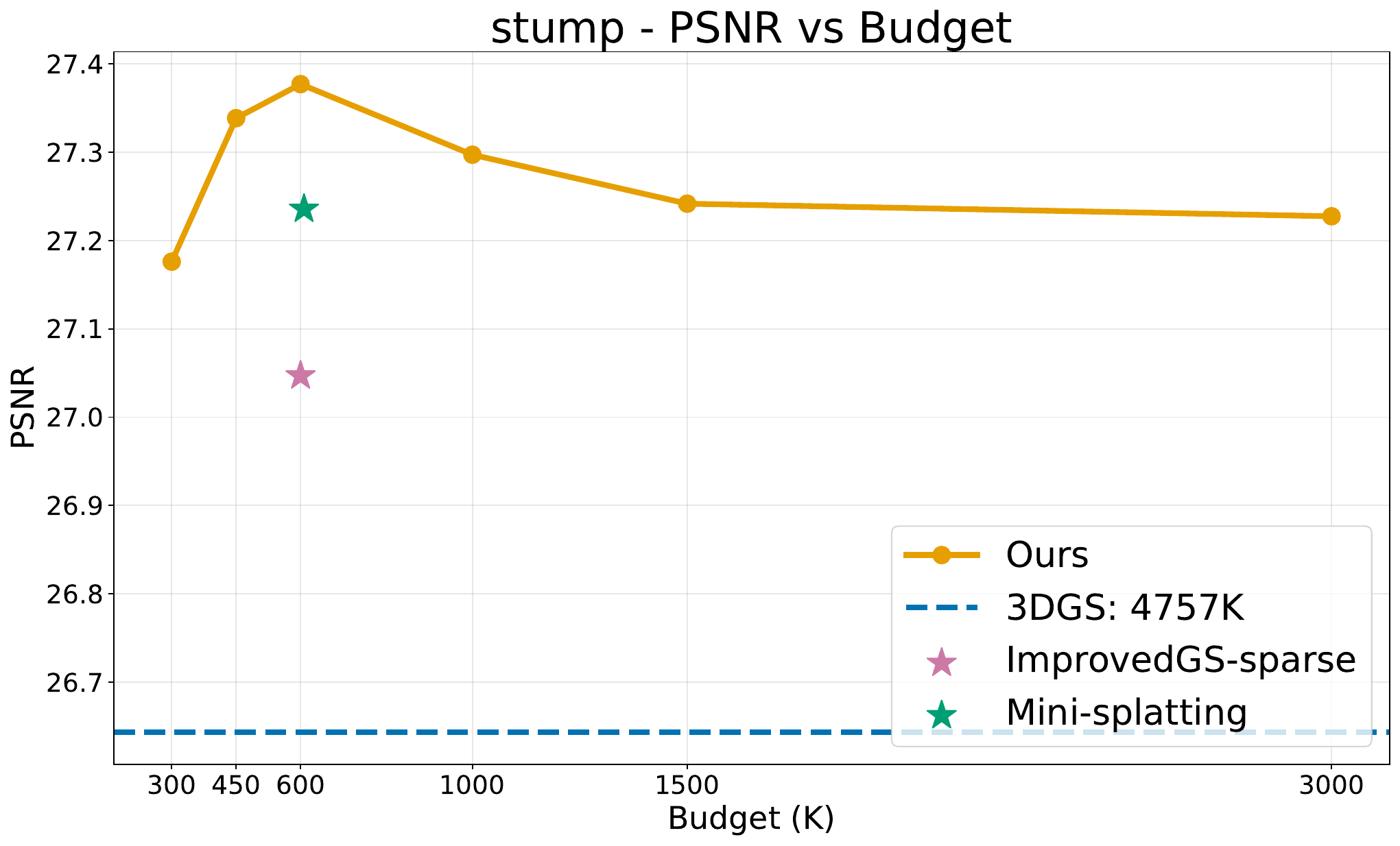}
\end{minipage}
\hfill
\begin{minipage}[b]{0.3\textwidth}
\centering
\includegraphics[width=\textwidth]{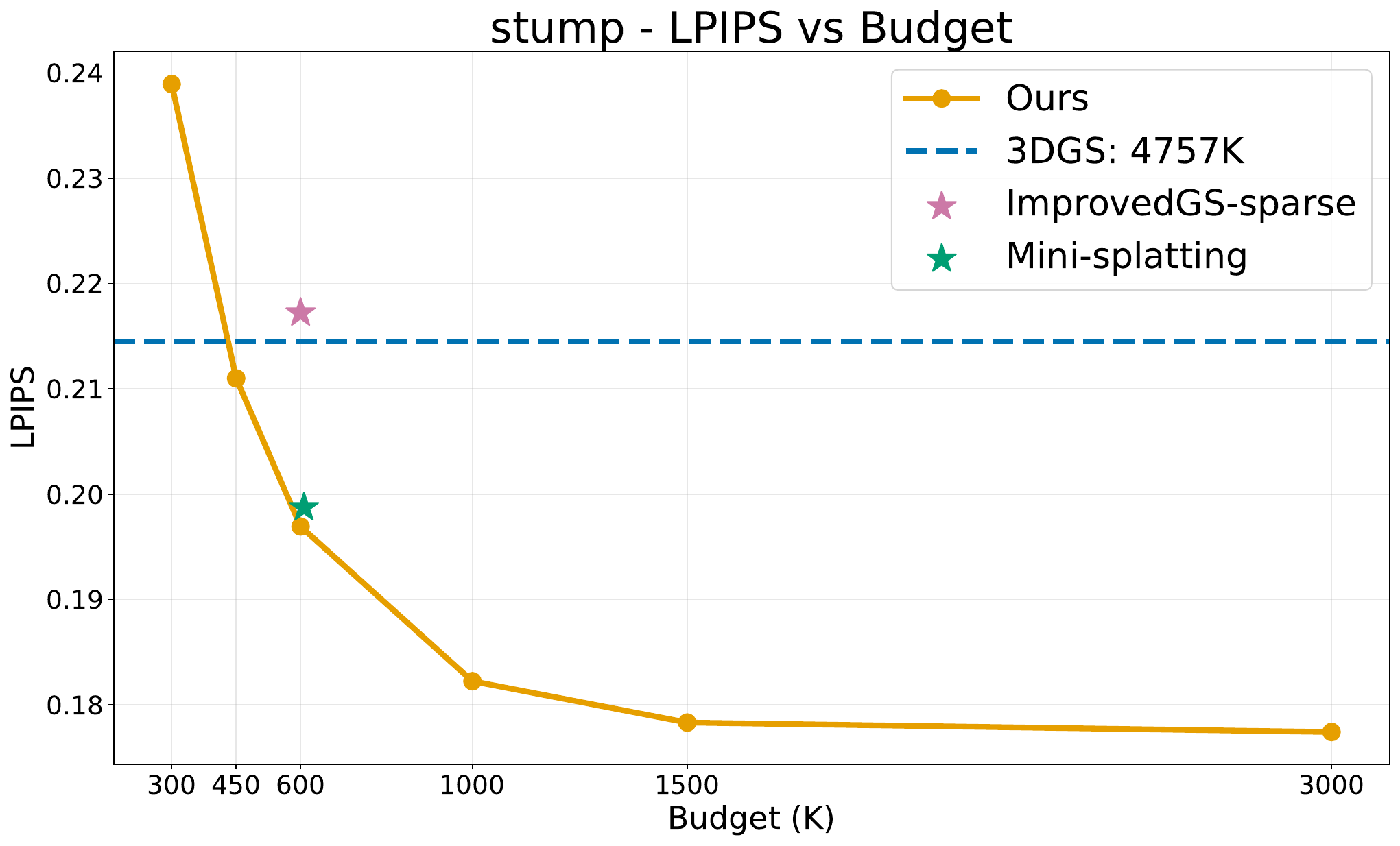}
\end{minipage}

\centering
\begin{minipage}[b]{0.3\textwidth}
\centering
\includegraphics[width=\textwidth]{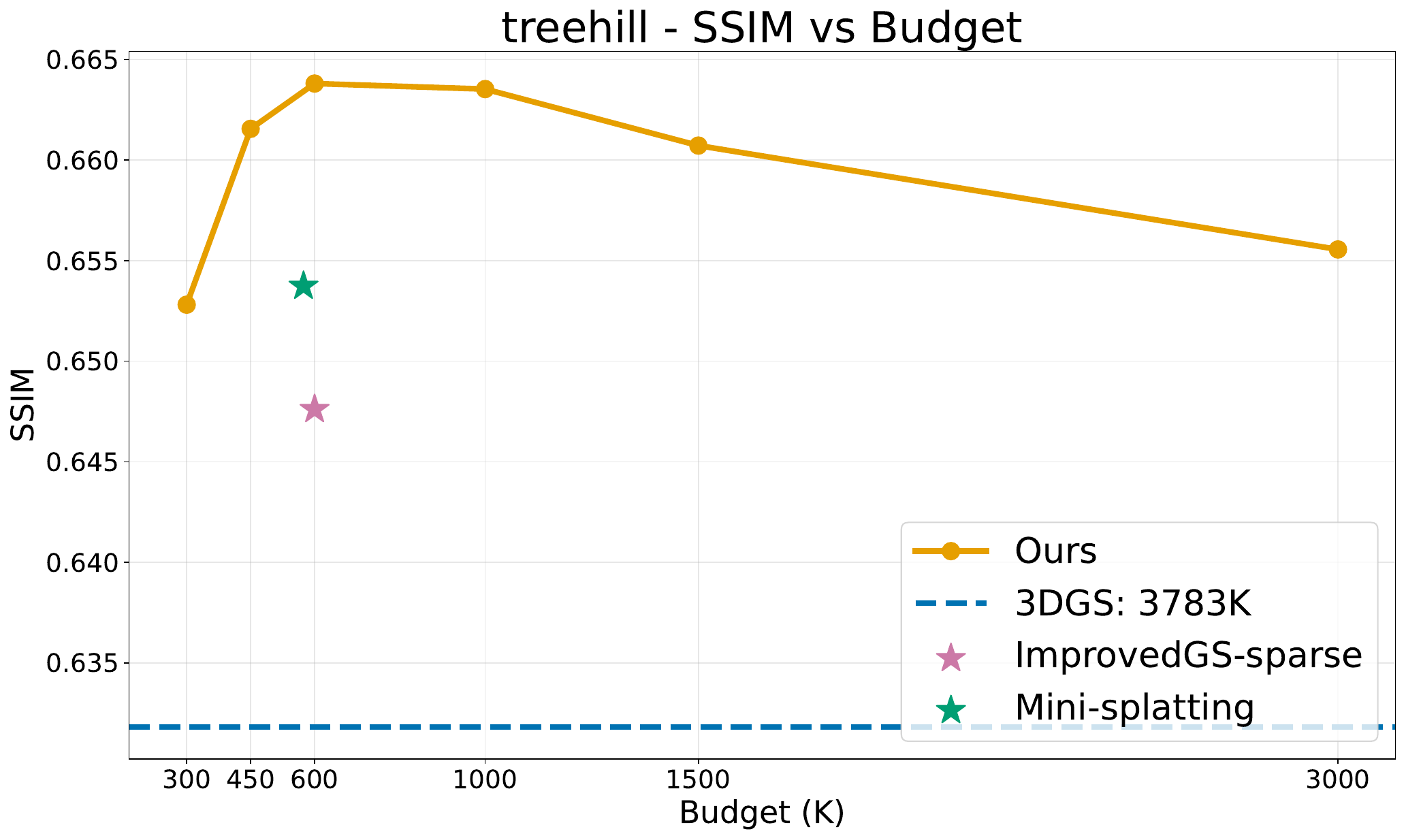}
\end{minipage}
\hfill
\begin{minipage}[b]{0.3\textwidth}
\centering
\includegraphics[width=\textwidth]{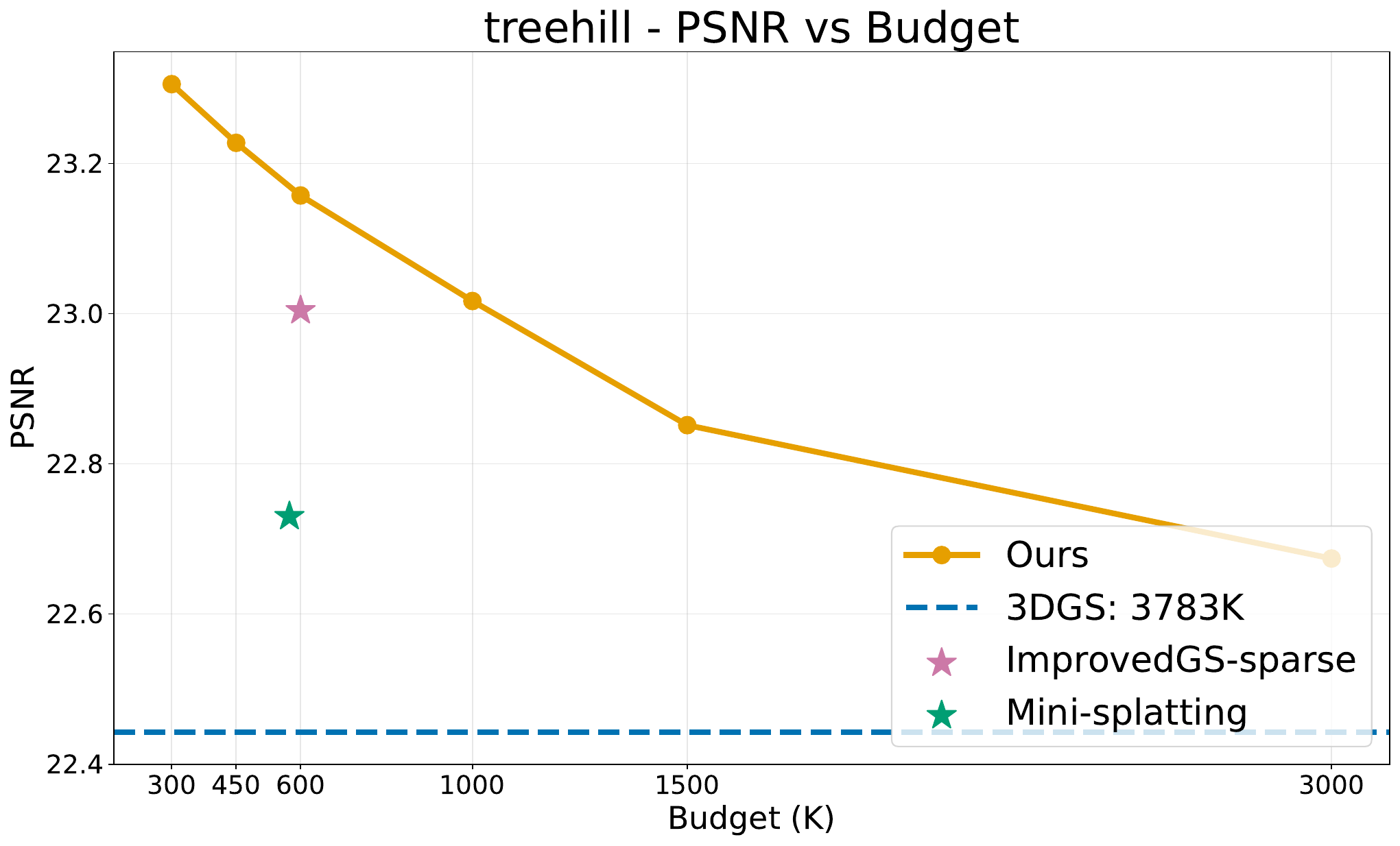}
\end{minipage}
\hfill
\begin{minipage}[b]{0.3\textwidth}
\centering
\includegraphics[width=\textwidth]{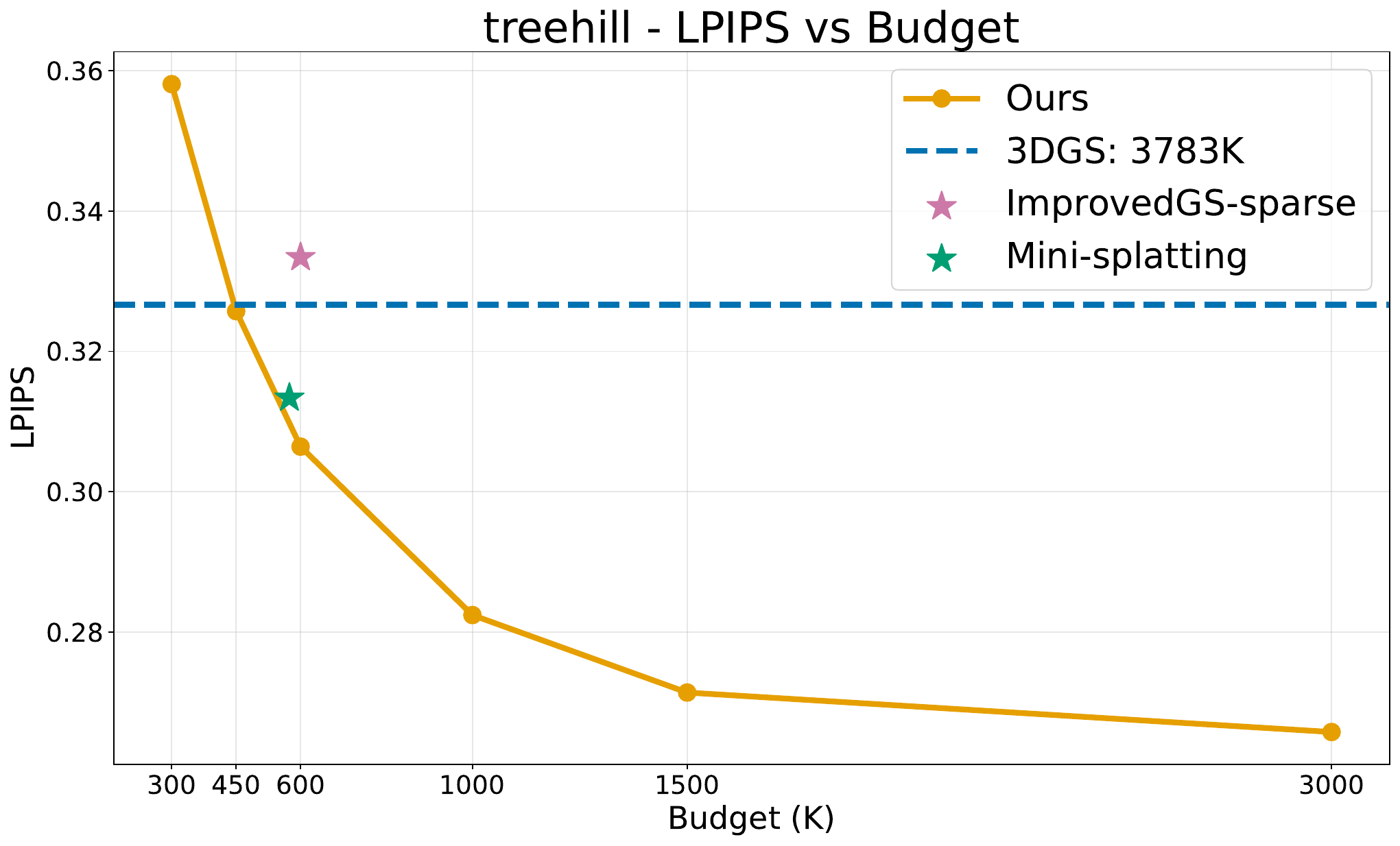}
\end{minipage}

\centering
\begin{minipage}[b]{0.3\textwidth}
\centering
\includegraphics[width=\textwidth]{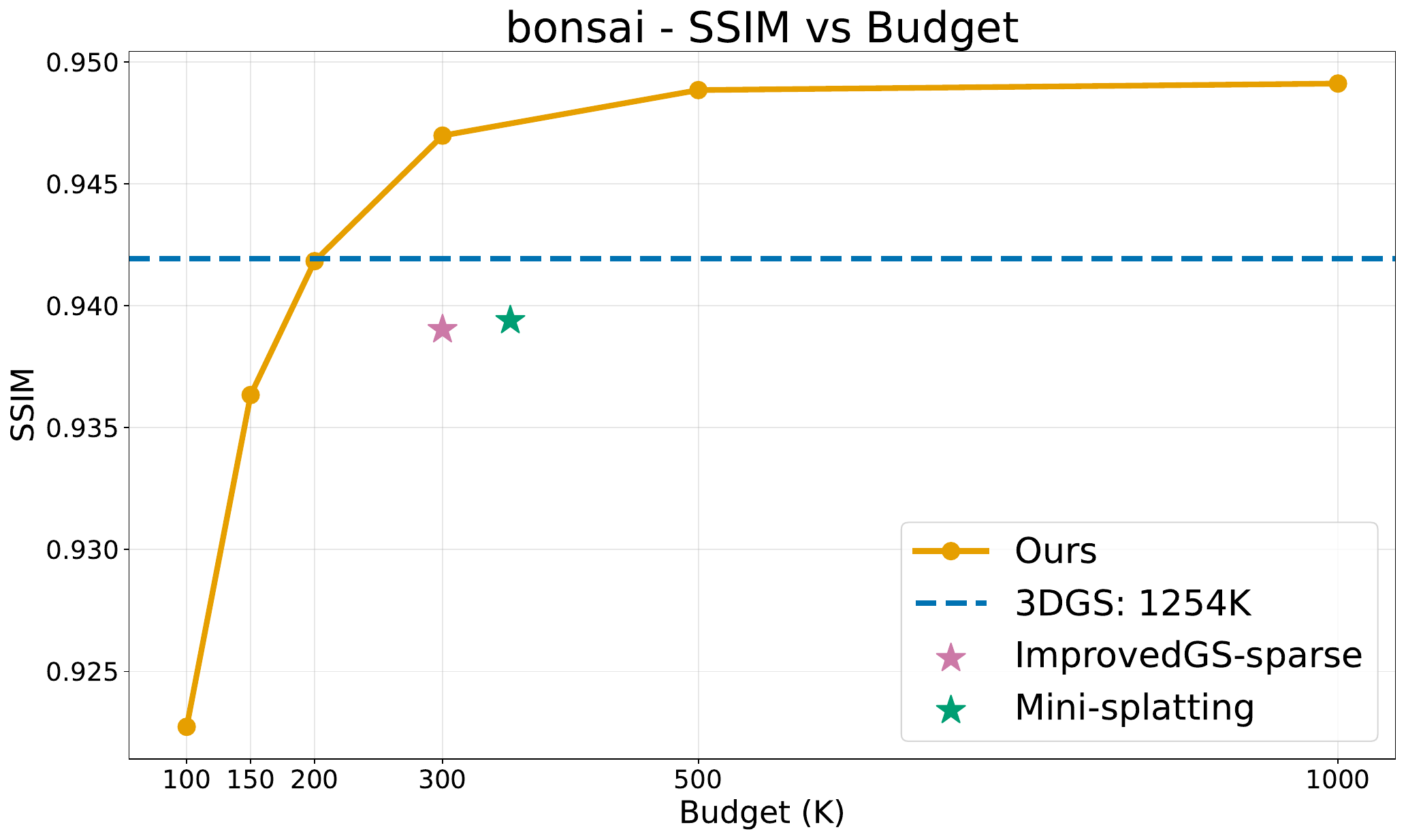}
\end{minipage}
\hfill
\begin{minipage}[b]{0.3\textwidth}
\centering
\includegraphics[width=\textwidth]{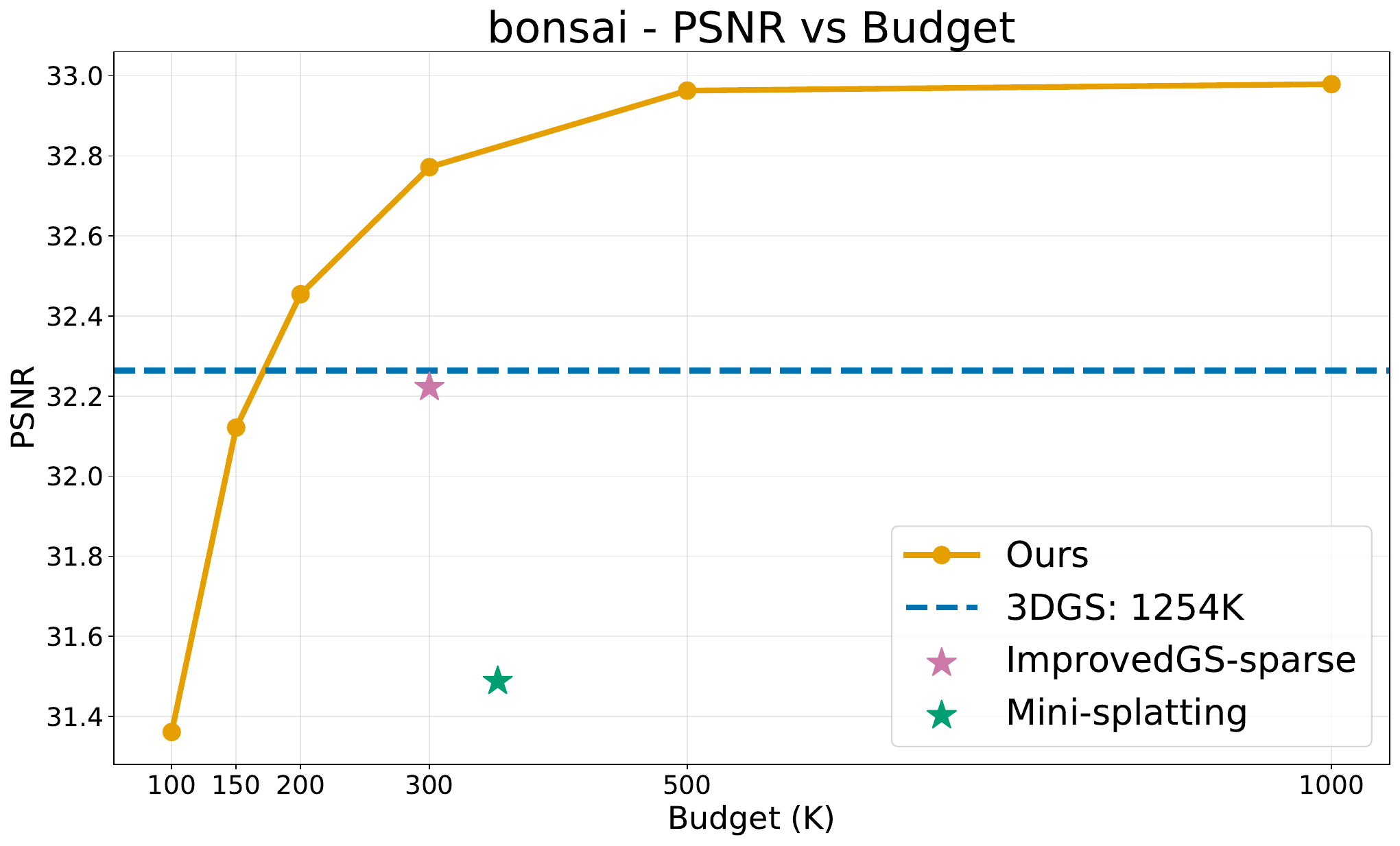}
\end{minipage}
\hfill
\begin{minipage}[b]{0.3\textwidth}
\centering
\includegraphics[width=\textwidth]{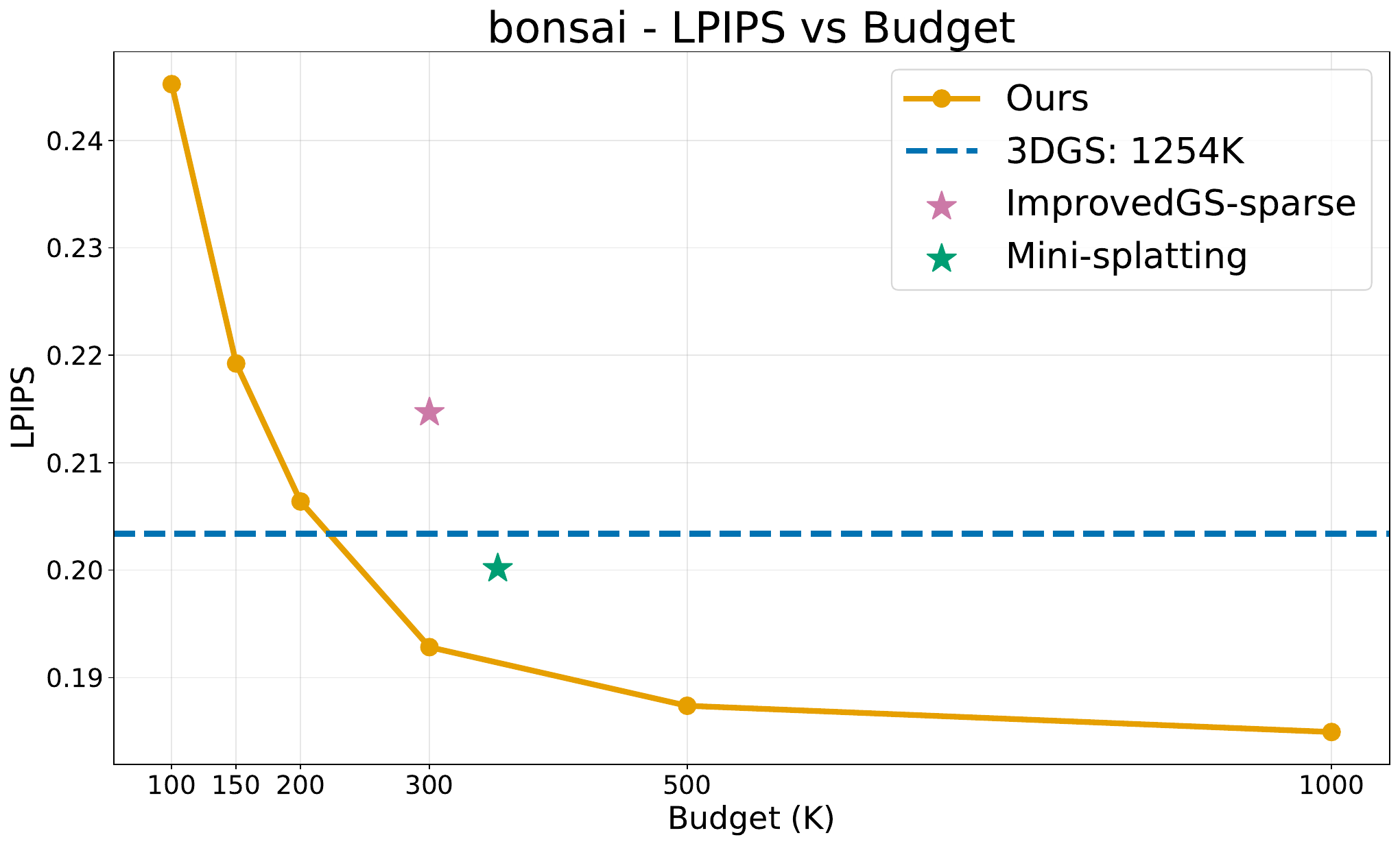}
\end{minipage}

\centering
\begin{minipage}[b]{0.3\textwidth}
\centering
\includegraphics[width=\textwidth]{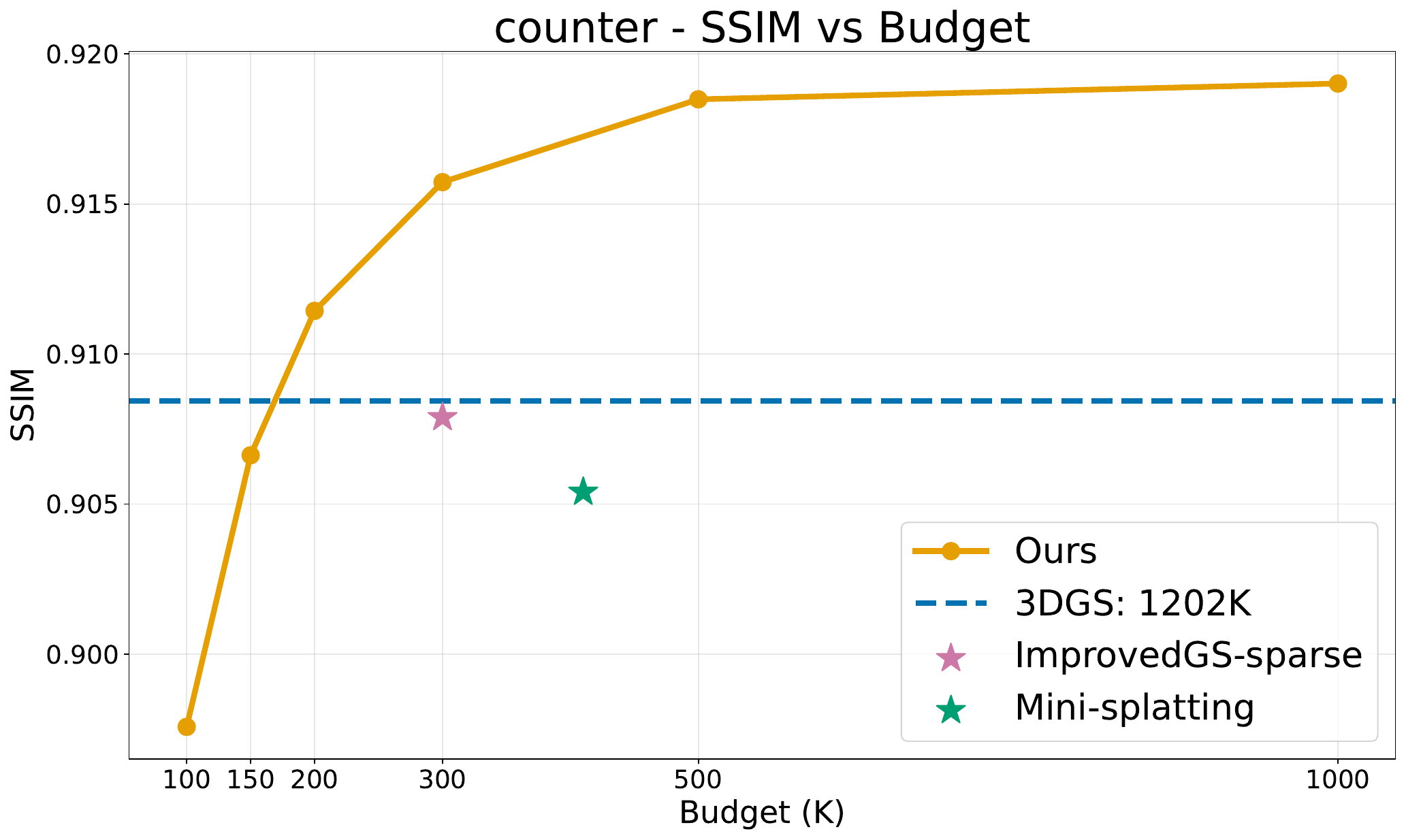}
\end{minipage}
\hfill
\begin{minipage}[b]{0.3\textwidth}
\centering
\includegraphics[width=\textwidth]{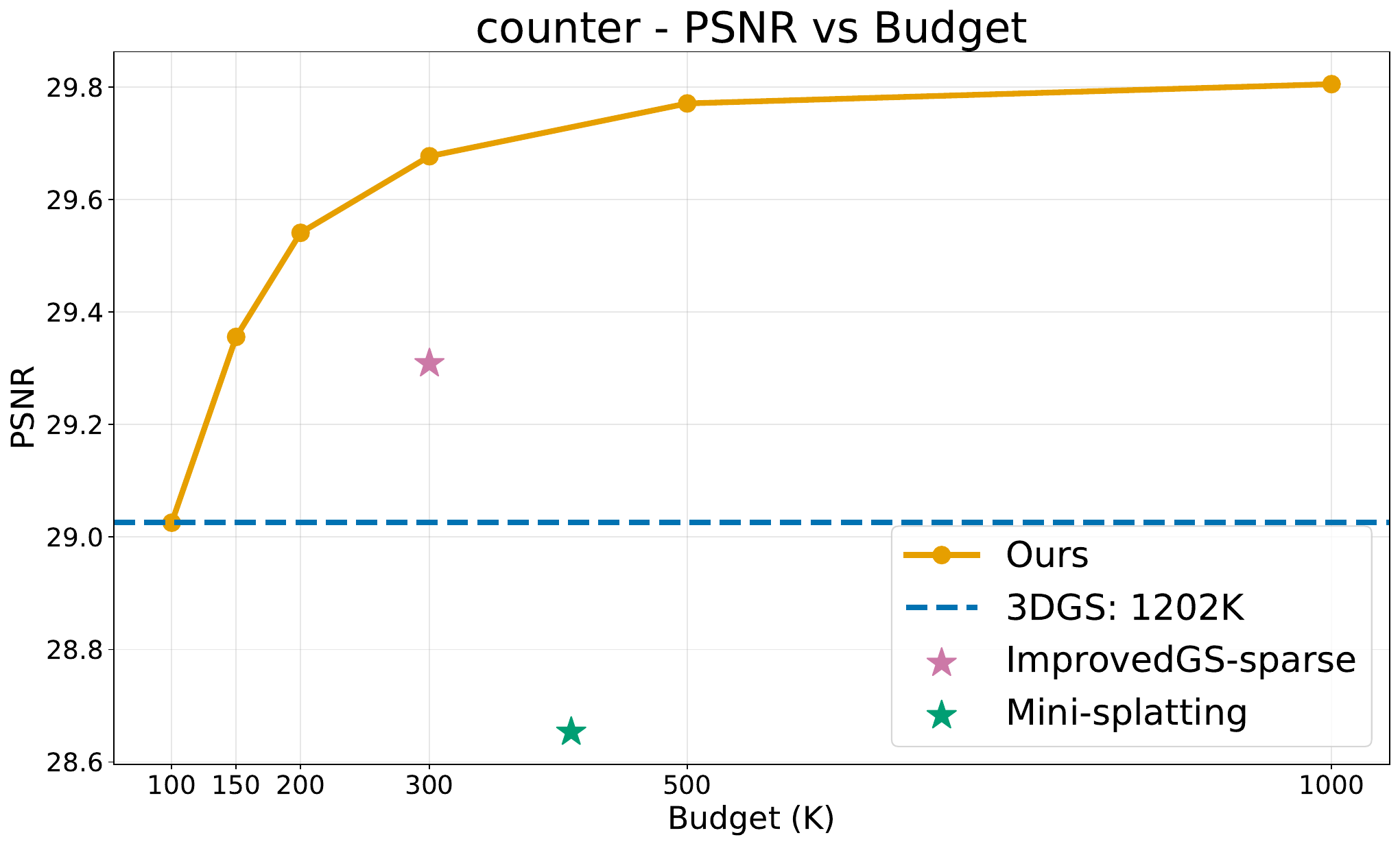}
\end{minipage}
\hfill
\begin{minipage}[b]{0.3\textwidth}
\centering
\includegraphics[width=\textwidth]{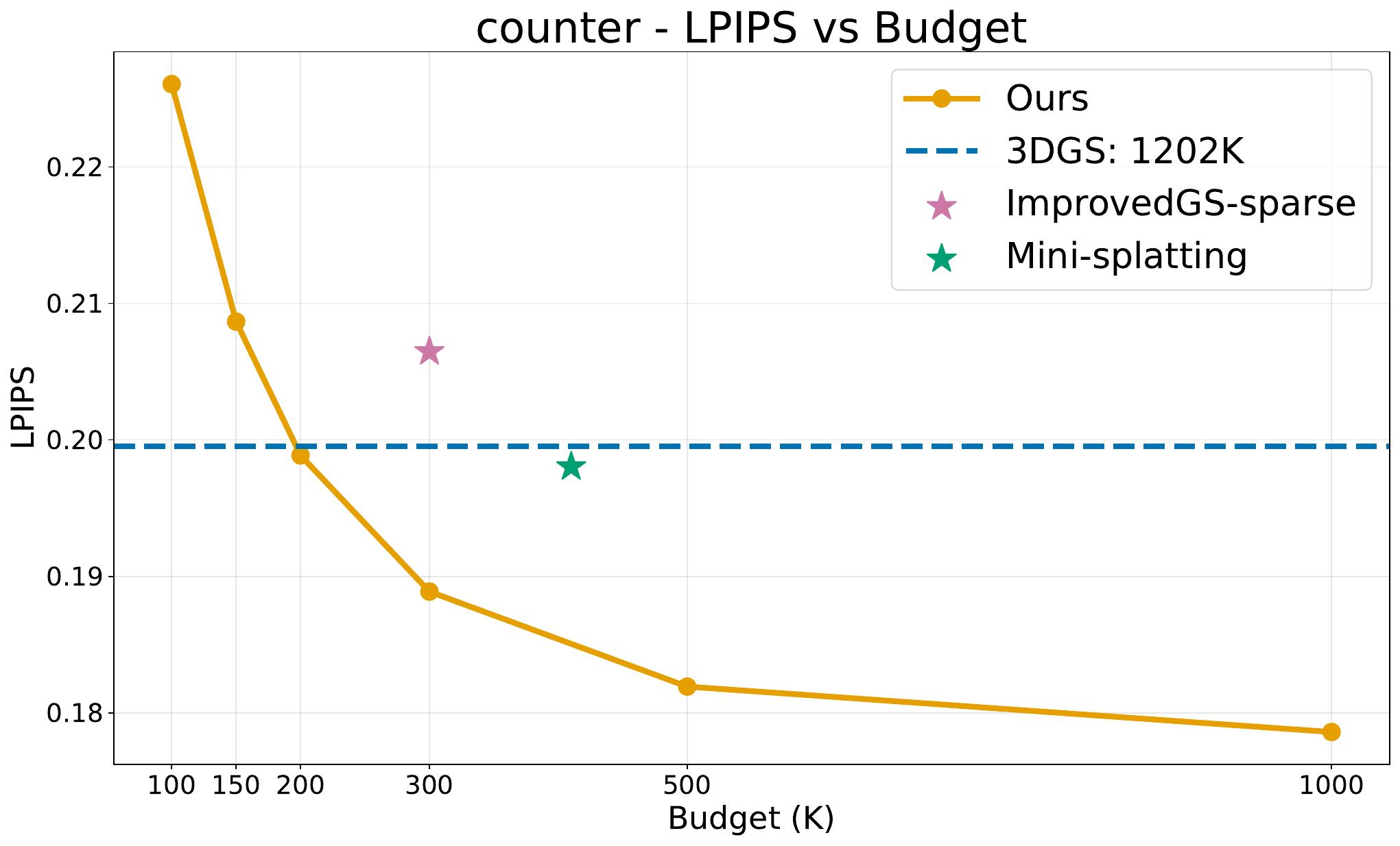}
\end{minipage}

\centering
\begin{minipage}[b]{0.3\textwidth}
\centering
\includegraphics[width=\textwidth]{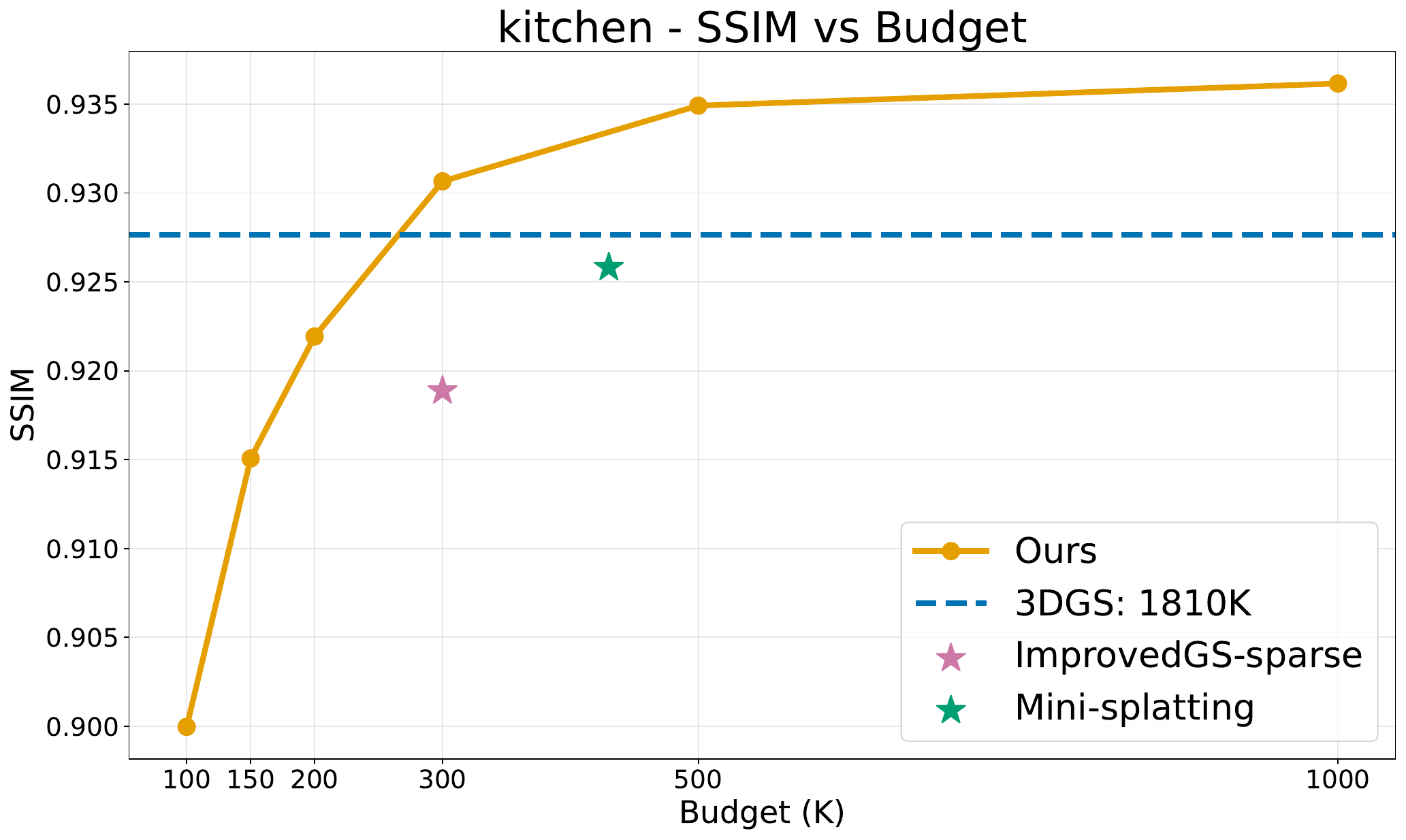}
\end{minipage}
\hfill
\begin{minipage}[b]{0.3\textwidth}
\centering
\includegraphics[width=\textwidth]{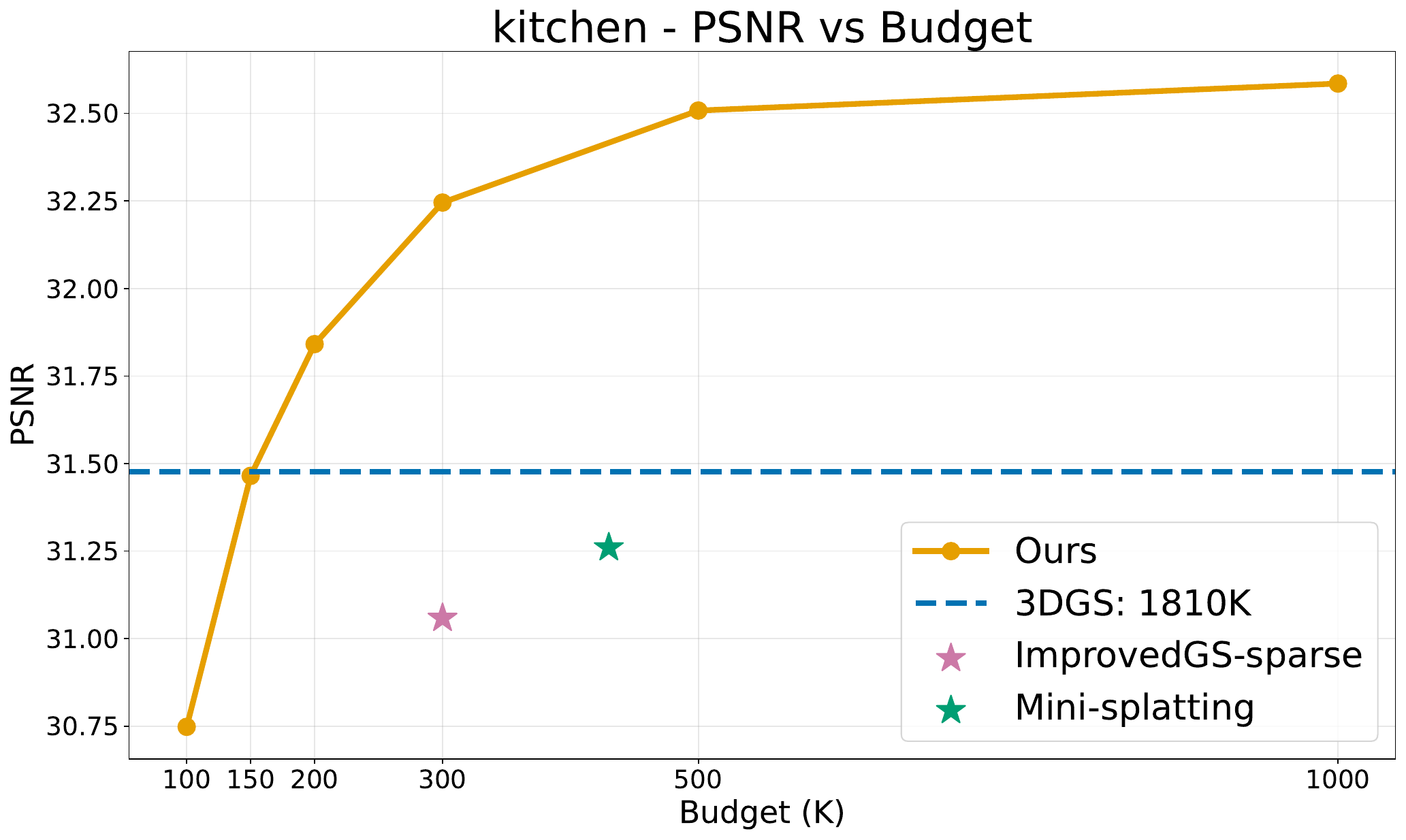}
\end{minipage}
\hfill
\begin{minipage}[b]{0.3\textwidth}
\centering
\includegraphics[width=\textwidth]{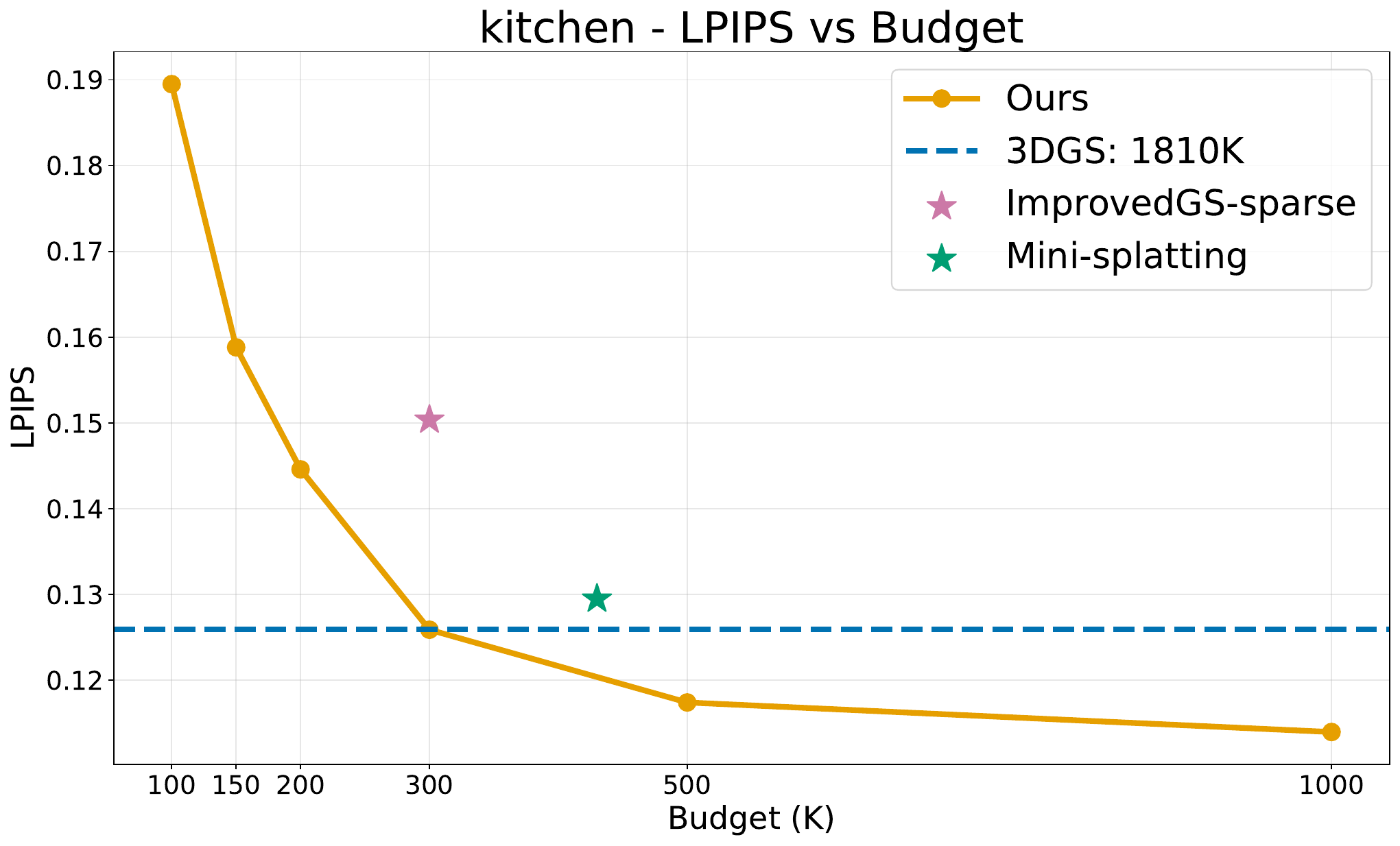}
\end{minipage}

\centering
\begin{minipage}[b]{0.3\textwidth}
\centering
\includegraphics[width=\textwidth]{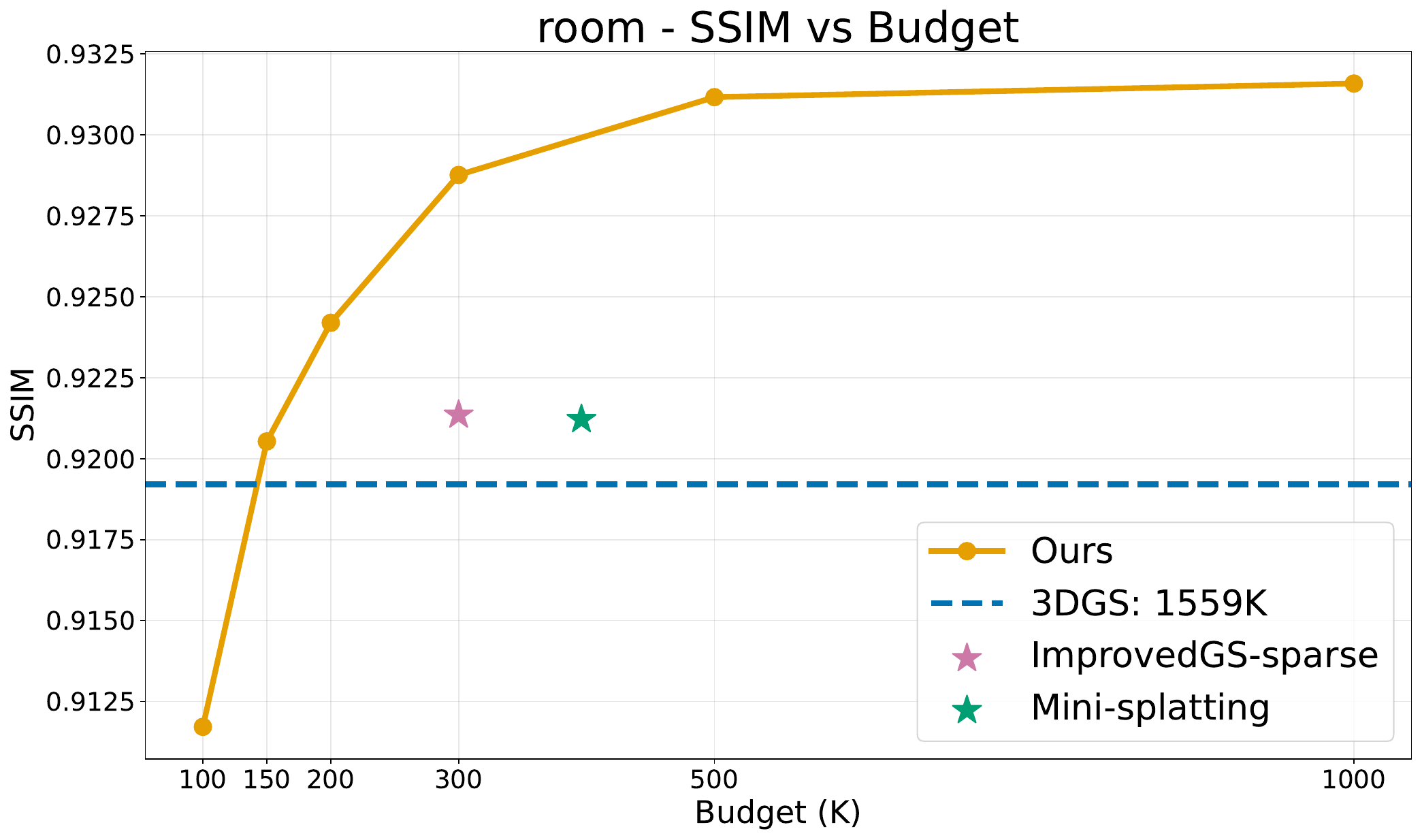}
\end{minipage}
\hfill
\begin{minipage}[b]{0.3\textwidth}
\centering
\includegraphics[width=\textwidth]{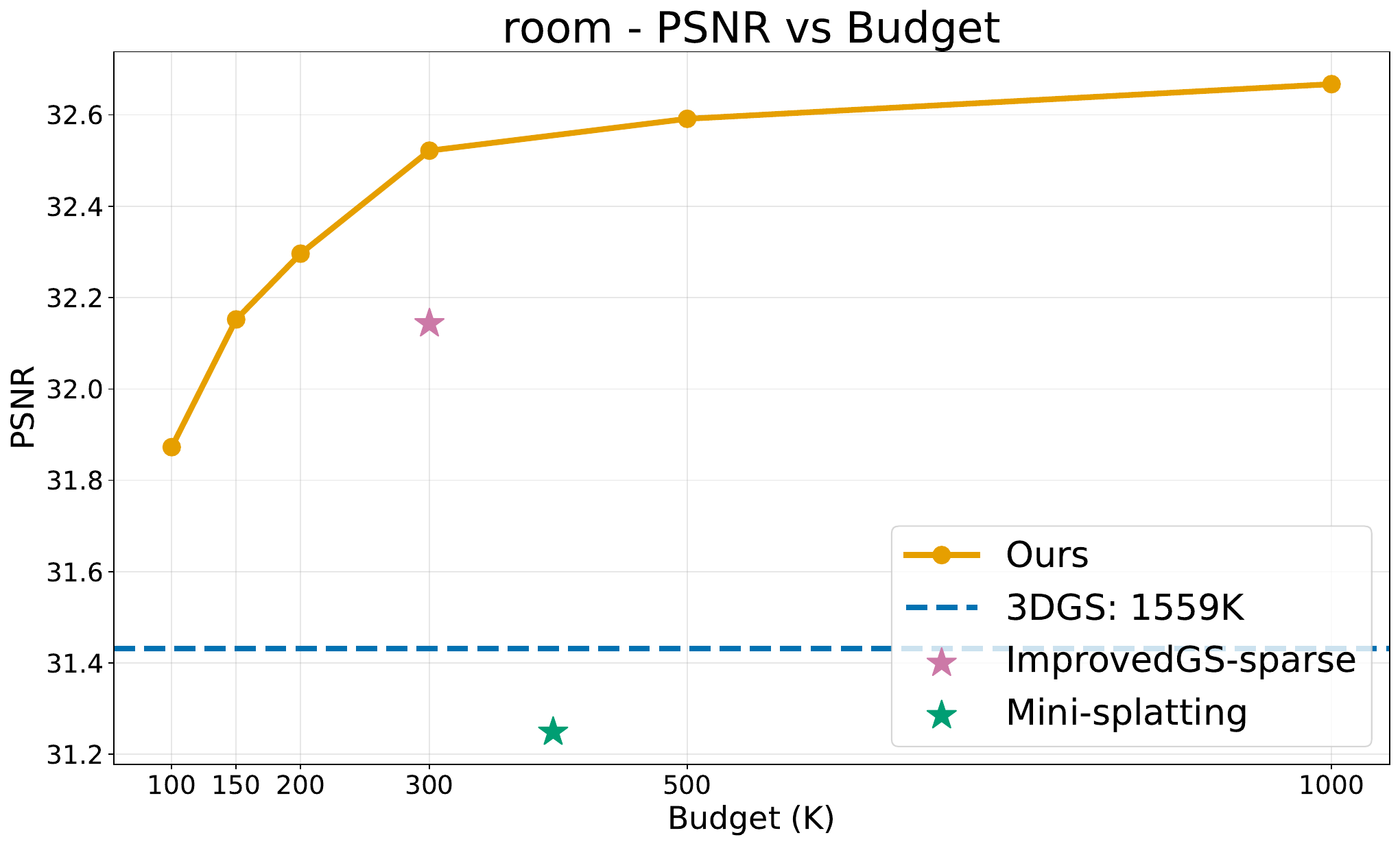}
\end{minipage}
\hfill
\begin{minipage}[b]{0.3\textwidth}
\centering
\includegraphics[width=\textwidth]{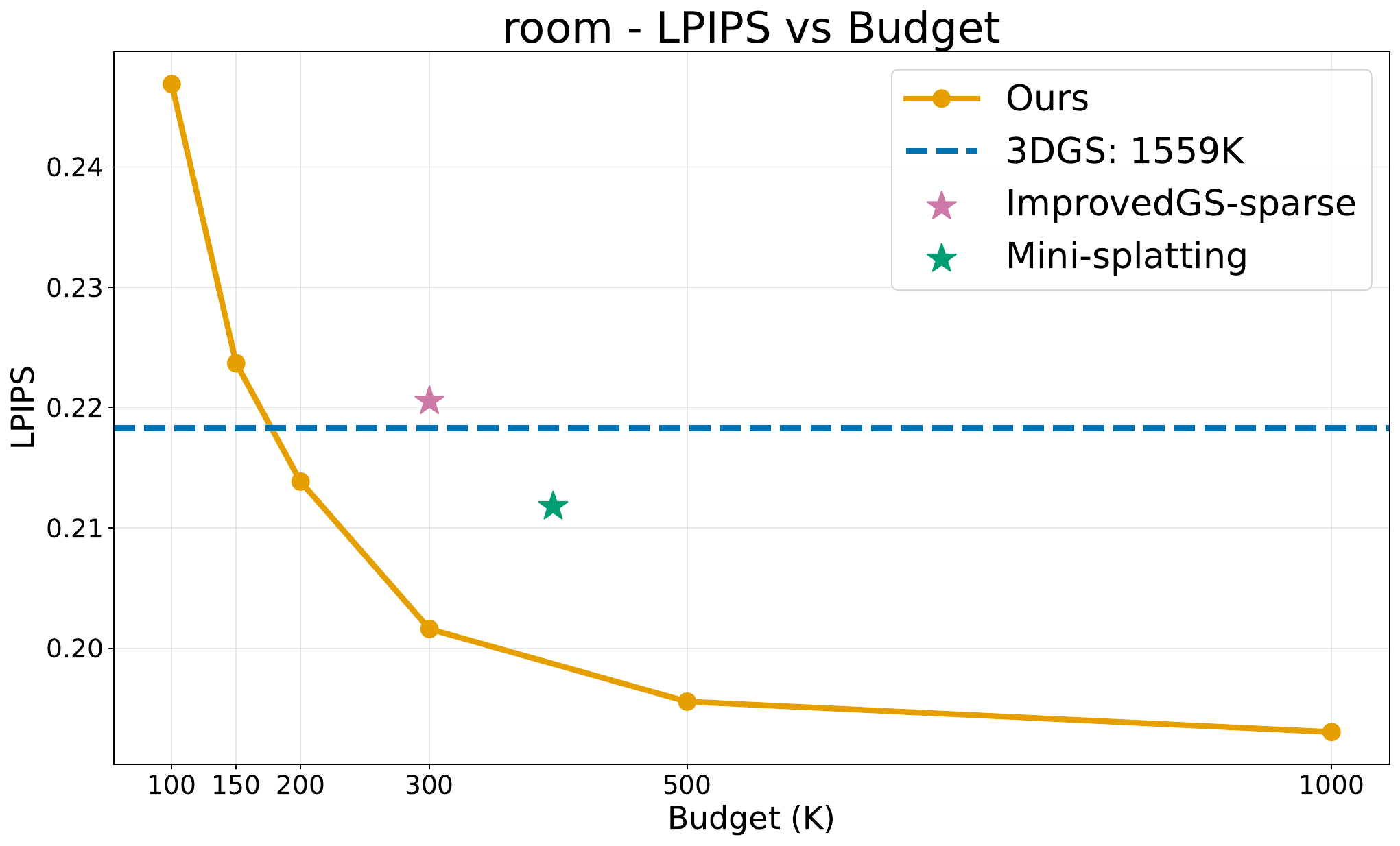}
\end{minipage}

\caption{Performance under varying Gaussian budgets of 6 scenes.}
\label{fig:score-budget3}
\end{figure*}

Figure~\ref{fig:score-budget2} and~\ref{fig:score-budget3} present the Performance Under Varying Gaussian Budgets across 8 scenes from the Mip-NeRF 360 dataset. 
Due to the inherent high volatility of the T \& T and DB datasets, similar comparisons are not conducted for these.

\subsection{More Qualitative Comparisons}
\begin{figure}[t]
    \centering
    \includegraphics[width=0.79\textwidth]{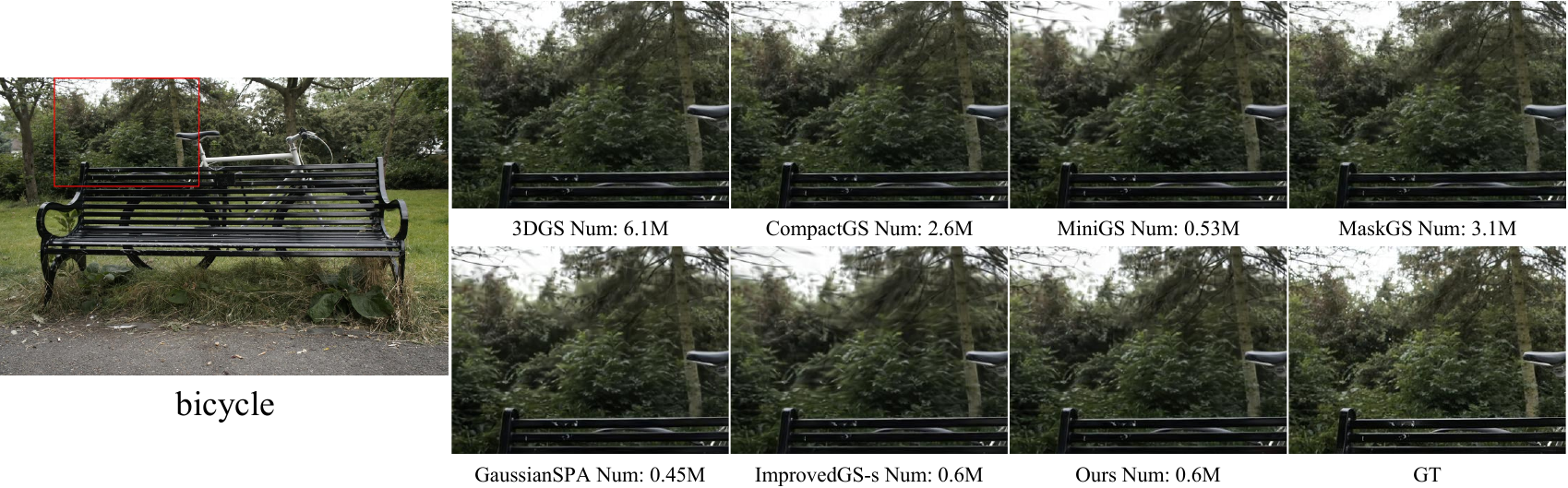} \\
    \includegraphics[width=0.79\textwidth]{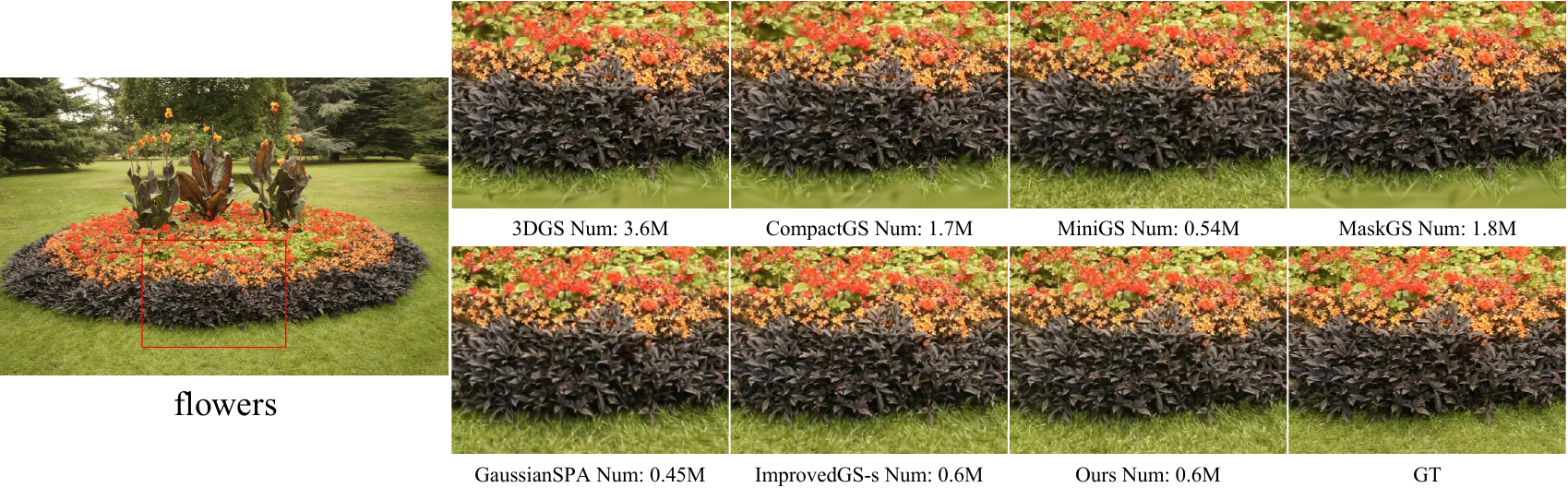} \\
    \includegraphics[width=0.79\textwidth]{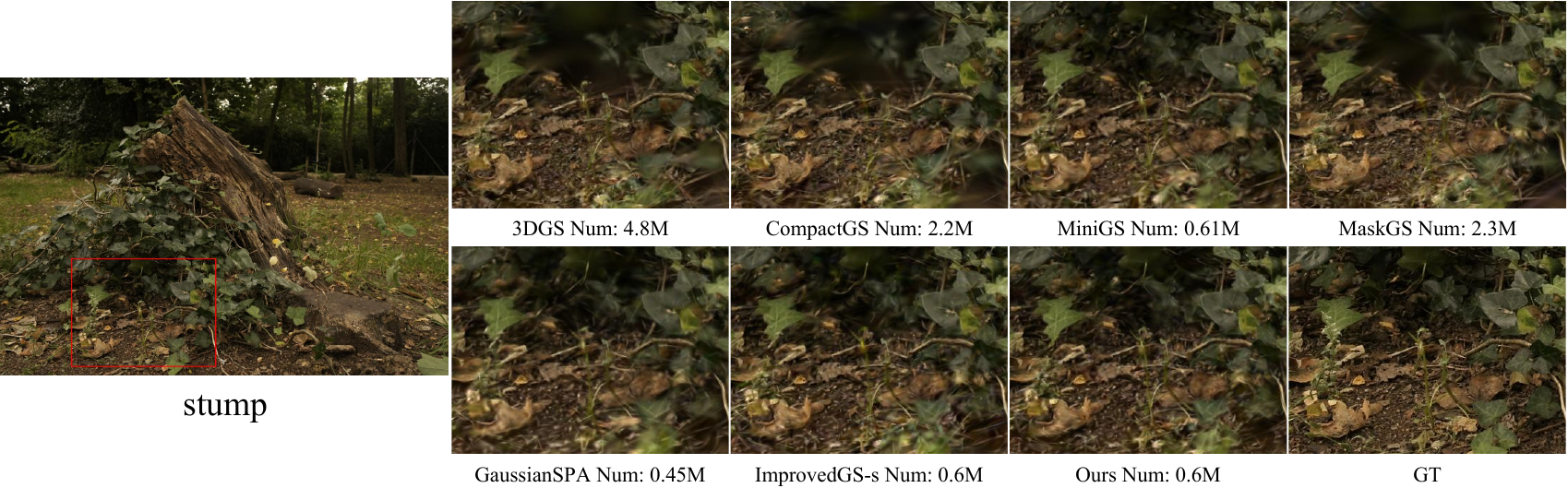} \\
    \includegraphics[width=0.79\textwidth]{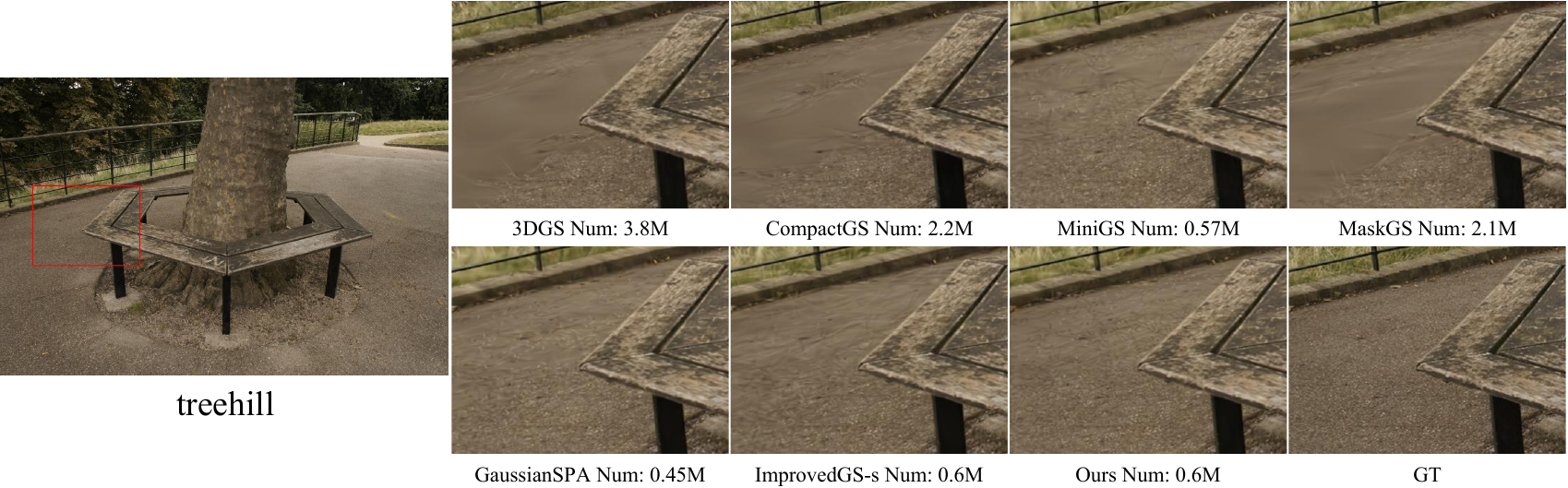} \\
    \includegraphics[width=0.79\textwidth]{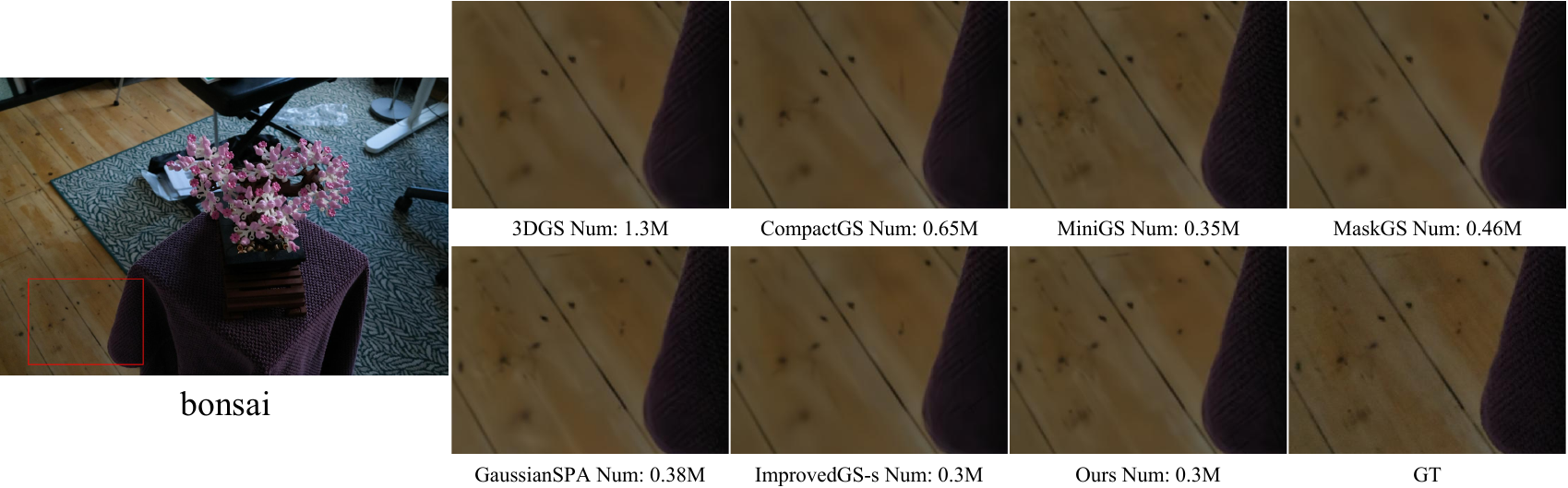} \\
    \caption{Qualitative comparison results among scenes bicycle, flowers, stump, treehill, bonsai.}
\label{fig:qualitative2}
\end{figure}

\begin{figure}[t]
    \centering
    \includegraphics[width=0.79\textwidth]{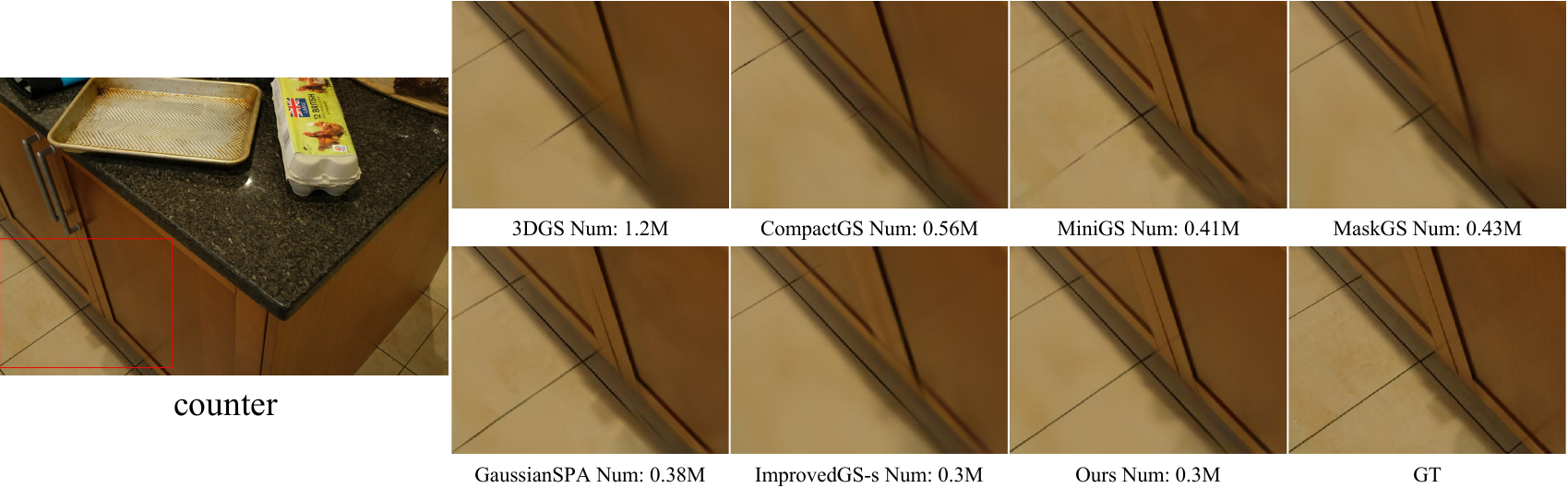} \\
    \includegraphics[width=0.79\textwidth]{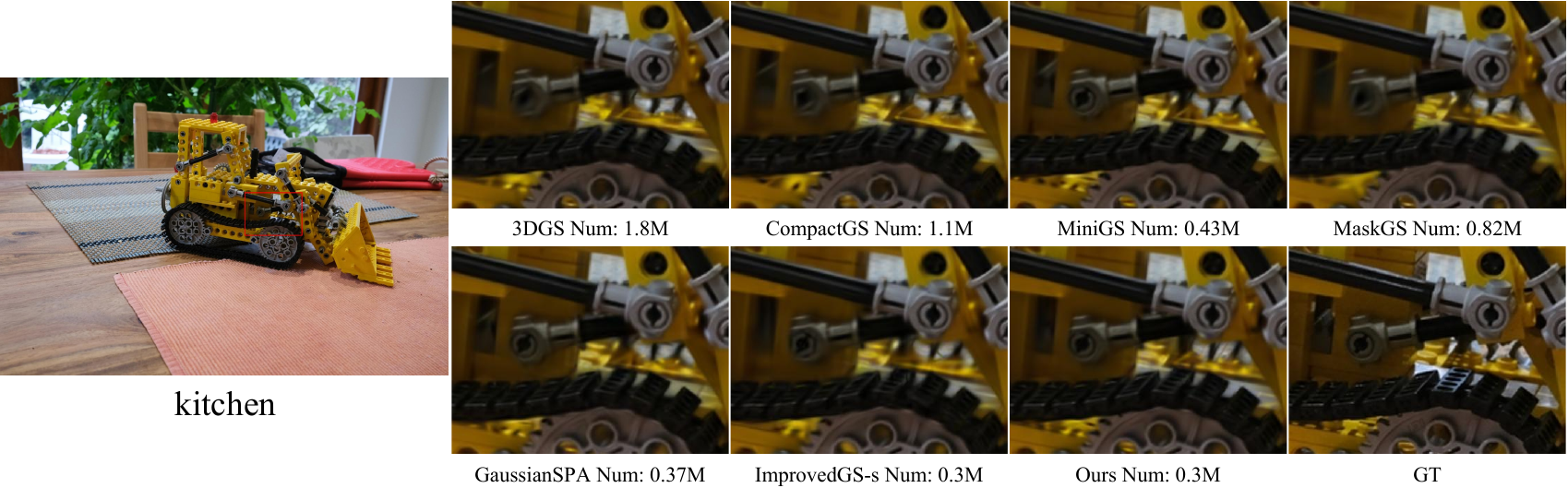} \\
    \includegraphics[width=0.79\textwidth]{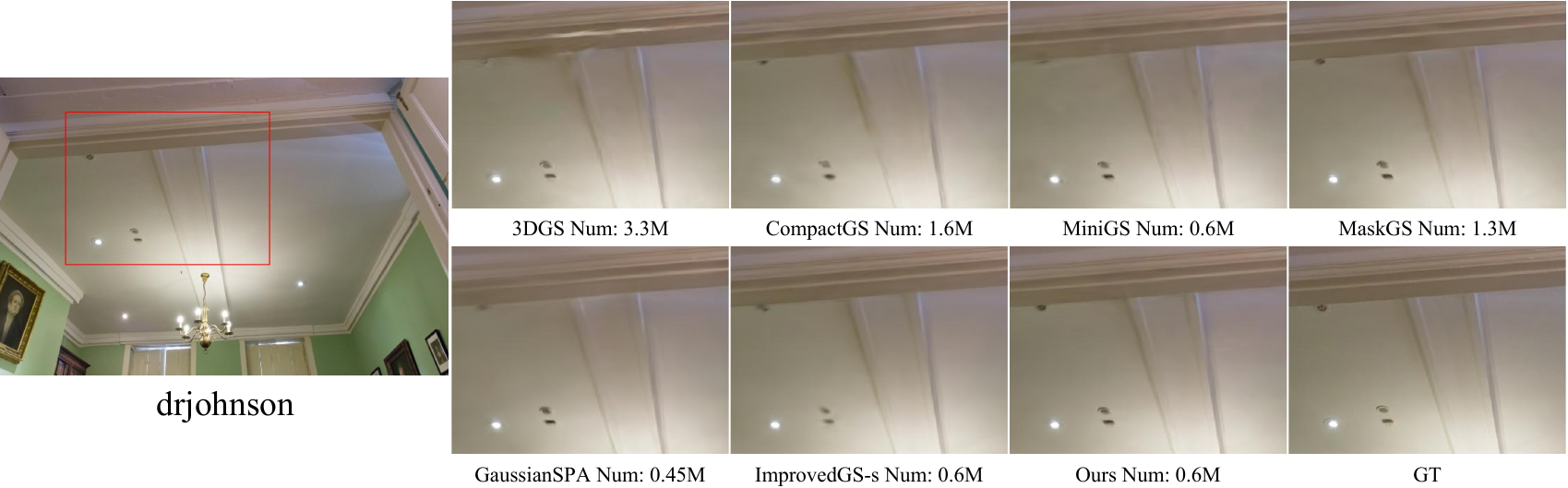} \\
    \includegraphics[width=0.79\textwidth]{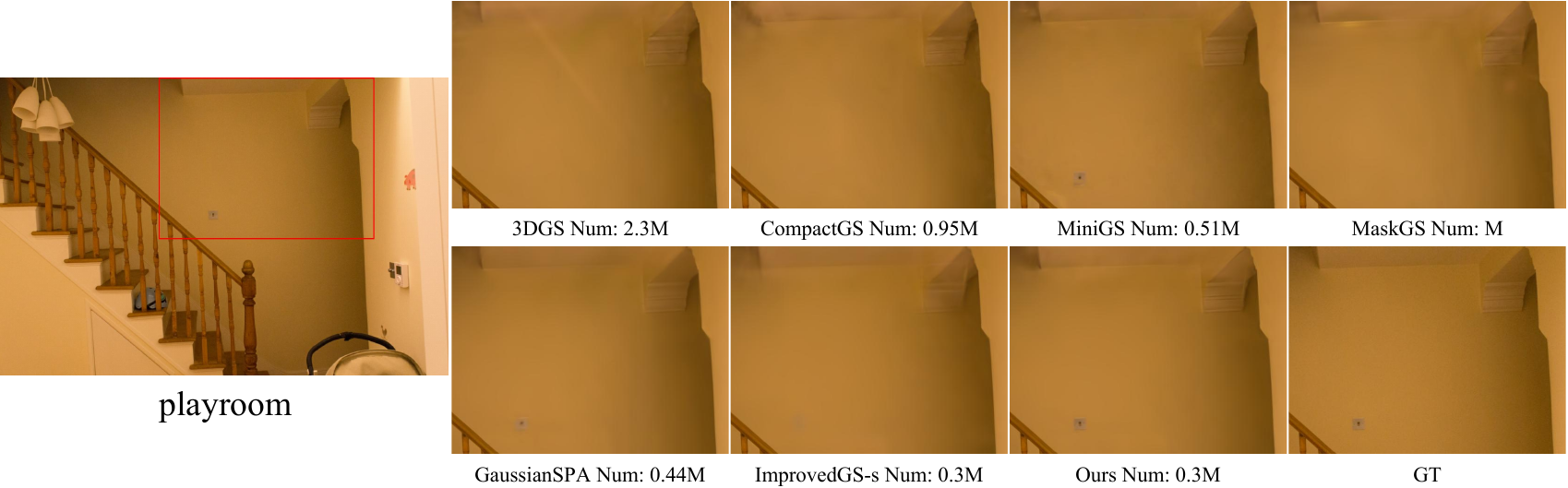} \\
    \includegraphics[width=0.79\textwidth]{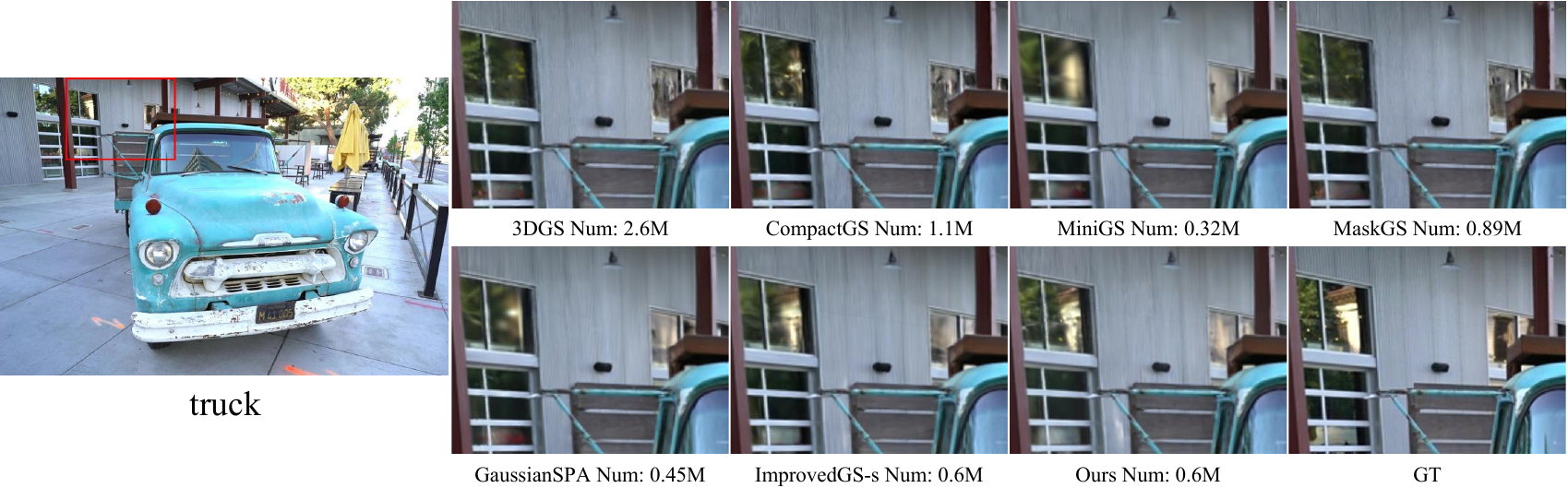} \\
    \caption{Qualitative comparison results among scenes counter, kitchen, drjohnson, playroom, truck.}
\label{fig:qualitative3}
\end{figure}

Figure~\ref{fig:qualitative2} and~\ref{fig:qualitative3} present qualitative comparisons of the remaining 10 scenes.
Our method achieves optimal rendering results across all scenes.

\clearpage
\subsection{Derivation of the Finite Prior}
This section derives the linear relation between the decay ratio and the current opacity under the finite prior.

Opacity is activated from the pre-activation variable $v$ by the sigmoid function:
\begin{equation}
    \alpha = S(v) = \frac{1}{1 + e^{-v}}.
\end{equation}

Ignoring rendering gradients, the update under environmental pressure is:
\begin{equation}
    v_{t+1} = v_t + \nabla v \cdot lr,
\end{equation}
which yields the decay ratio:
\begin{equation}
    R_o^{t+1} = \frac{S(v_t) - S(v_t + \nabla v \cdot lr)}{S(v_t)}.
\end{equation}

Let $\Delta v = \nabla v \cdot lr$ and assume it is small, i.e., $|\Delta v| \ll 1$. A first-order Taylor expansion of $S(v_t + \Delta v)$ at $v_t$ gives:
\begin{equation}
    S(v_t + \Delta v) \approx S(v_t) + S'(v_t)\Delta v.
\end{equation}

Since the derivative of the sigmoid is
\begin{equation}
    S'(v) = S(v)[1 - S(v)] = \alpha(1 - \alpha),
\end{equation}
we obtain
\begin{equation}
    S(v_t + \Delta v) \approx \alpha_t + \alpha_t(1-\alpha_t)\Delta v.
\end{equation}

Substituting into the decay ratio:
\begin{equation}
\begin{aligned}
    R_o^{t+1}
    &= \frac{\alpha_t - S(v_t + \Delta v)}{\alpha_t} \\
    &\approx -(1 - \alpha_t)\Delta v.
\end{aligned}
\end{equation}

Environmental pressure enforces $\Delta v < 0$, so we take the magnitude:
\begin{equation}
    R_o^{t+1} \approx (1 - \alpha_t)\,|\Delta v|
    = (1 - \alpha)\,|\nabla v \cdot lr|.
\end{equation}

When $|\Delta v| \to 0$, the approximation error is $\mathcal{O}(|\Delta v|^2)$, and the first-order expression becomes accurate.

Thus, the decay ratio is linearly proportional to $(1-\alpha)$:
\begin{itemize}
    \item As $\alpha \to 1$, the decay ratio vanishes, preserving consistently high-fitness Gaussians;
    \item As $\alpha \to 0$, the decay ratio approaches its maximum $|\nabla v \cdot lr|$, accelerating the removal of low-fitness individuals.
\end{itemize}

This finite prior therefore accelerates convergence while maintaining the fairness and adaptivity of natural selection.

\subsection{Pruning based on 3DGS}

Our method employs Improved-GS as the densification strategy to achieve the best compact 3DGS rendering quality, but this does not imply that our approach relies on specific prior work. 
Table~\ref{tab:quantitative_3dgs} also presents the results of our method on 3DGS, with the budget uniformly set to 1/4 of the peak densification budget in 3DGS. 
Even within the 3DGS framework, our method achieves highly competitive pruning performance.

\begin{table}[ht]
	\centering
		\begin{tabular}{l|cccc}
			Dataset & \multicolumn{4}{c|}{Mip-NeRF360}\\
			Method|Metric
			& $SSIM^\uparrow$ & $PSNR^\uparrow$  & $LPIPS^\downarrow$  & $Num^\downarrow$\\
			\hline
			3DGS & 0.816  & 27.50  & 0.216  & 3320453\\
            \hline
            Compact-3DGS & 0.807  & 27.33  & 0.227  & 1516172\\
            MaskGS & 0.815  & 27.43  & 0.218  & 1582926\\
            \hline
            Ours(3DGS) & 0.815  & 27.45  & 0.222  & 830000\\
        \end{tabular}
	\caption{Quantitative results based on 3DGS densification method.}
	\label{tab:quantitative_3dgs}
\end{table}

\clearpage
\subsection{Opacity Learning Rate Scaling}

\textbf{Necessity.}
To complete the natural selection process within a limited number of training iterations, a sufficiently strong regularization gradient is required. 
If the original opacity learning rate is maintained, the selection outcome is heavily influenced by the \emph{initial} opacity distribution, since the final opacity can be decomposed as:
\[
\alpha_{\text{final}} = \alpha_{\text{init}} + \text{cumulative recovery} - \text{cumulative decay}.
\]
Increasing the opacity learning rate during natural selection proportionally amplifies both the recovery and decay terms, thereby increasing the relative influence of dynamic optimization gradients and reducing the dependence on initial opacity.

\begin{table}[ht]
	\centering
		\begin{tabular}{l|cccc}
			Factors
			& $SSIM^\uparrow$ & $PSNR^\uparrow$  & $LPIPS^\downarrow$\\
			\hline
			$1\times$ & 0.786  & 25.71  & 0.227\\
            $2\times$ & 0.790  & 25.72  & 0.217\\
            $3\times$ & 0.792  & 25.80  & 0.214\\
            $4\times$ & 0.793  & 25.80  & 0.214\\
            $5\times$ & 0.792  & 25.76  & 0.213\\
        \end{tabular}
	\caption{The impact of different opacity learning rate scaling factors on the rendering quality of the bicycle scene.}
	\label{tab:opacity-lr}
\end{table}

On the \textit{bicycle} scene, we evaluate different scaling factors (Table~\ref{tab:opacity-lr}). A scale of $4\times$ already approaches the upper quality bound. 
However, excessively large learning rates also increase per-step decay, which may violate the assumption that gradients are approximately zero under the finite prior. 
Thus, we adopt a scaling factor of $4\times$ as a balanced choice.
After natural selection concludes, we allow the scene an additional 1,000 iterations for opacity recovery, after which the opacity learning rate will revert to its original value.

\subsection{Learning Rate of the Regularization Gradient Field}

\begin{table}[ht]
	\centering
		\begin{tabular}{l|cccc}
			lr
			& $SSIM^\uparrow$ & $PSNR^\uparrow$  & $LPIPS^\downarrow$\\
			\hline
			$0.002$ & 0.773  & 25.51  & 0.239\\
            $0.0025$ & 0.792  & 25.78  & 0.213\\
            $0.003$ & 0.793  & 25.78  & 0.213\\
            $0.004$ & 0.791  & 25.78  & 0.217\\
            $0.006$ & 0.787  & 25.72  & 0.226\\
            $0.008$ & 0.784  & 25.71  & 0.232\\
        \end{tabular}
	\caption{The impact of regularization gradient field learning rate on the rendering quality of the bicycle scene.}
	\label{tab:regularization-lr}
\end{table}

\begin{figure}[t]
    \centering
    \includegraphics[width=0.8\textwidth]{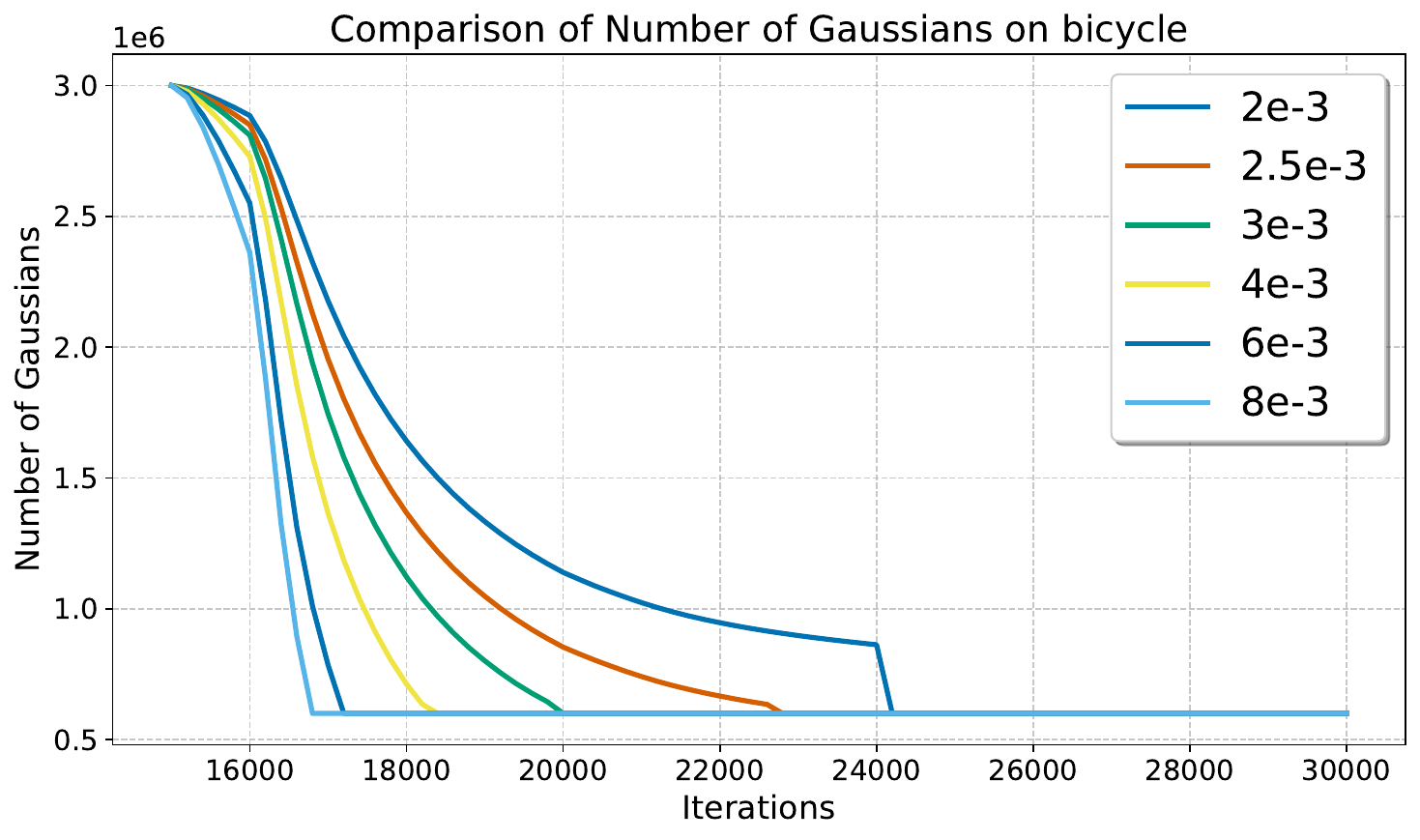}
    \caption{In the bicycle scene, pruning curves corresponding to different regularization learning rates. }
    \label{fig:num_lr}
\end{figure}

\textbf{Effect of the learning rate.}
The learning rate of the regularization gradient field directly controls the rate at which natural selection proceeds. 
After selection completes, a short period of fine-tuning is still required.
Empirically, allowing natural selection to run for approximately \textbf{5K–8K} iterations yields the highest pruning quality.

We set the latest possible termination point at \textbf{23K} iterations. 
If the budget is not met by that point, a one-shot pruning based on opacity is applied.
In the \textit{bicycle} scene, six learning rates are tested (Table~\ref{tab:regularization-lr}, Fig.~\ref{fig:num_lr}). 
The second-smallest learning rate—used in our main experiments—enables natural selection to finish just before the latest allowed iteration.

Results show:
\begin{itemize}
    \item Natural selection lasting 5K–8K iterations consistently achieves the best rendering quality.
    \item Insufficient selection time degrades pruning quality, though moderately.
    \item If the process does not finish on time and must fall back to one-shot pruning, rendering quality drops substantially due to the harshness and instability of one-shot removal.
\end{itemize}

\textbf{Learning rate selection.}
Across scenes, we observe that \textbf{higher redundancy} corresponds to \textbf{lower required learning rates}. Redundancy depends on scene complexity (resolution, detail density) and the extent to which high Gaussian counts actually improve rendering quality. 
Indoor scenes usually exhibit lower complexity and redundancy, while higher-resolution scenes contain richer details and thus lower redundancy.

\subsection{Automated Learning Rate}
By tracking the opacity change curve of Gaussians ranked at the final budget under the optimal learning rate, we observe that the overall trend is linear. 
Consequently, we can dynamically adjust the learning rate by comparing the opacity of the current target Gaussian with the preset curve. 
This version was completed after the paper submission and has been included in the code provided in Supplementary Material. 
Additionally, this version sets the densification budget to three times the final budget, meaning that for each scene, only the final budget parameter needs to be specified. 
This parameter can be easily automated by adjusting the gradient threshold for densification. 
However, manual control of the final budget remains a desirable feature, so we have chosen to retain it.

The automated learning rate version achieves rendering quality nearly identical to that of manual parameter tuning.
However, since the main text version still employs manual parameter tuning, the results provided in the appendix are also based on manually tuned parameters.
\subsection{FPS Evaluation}

\begin{table}[ht]
	\centering
    \scalebox{0.72}{
	\begin{tabular}{l|c|c|c|c|c|c|c}
		Methods & ImprovedGS-s & SPA	& MaskGS & Opacity Pruning	& Render Pruning & Edge Pruning	& Natural Selection(Ours)\\
        \hline
        FPS & 153 & 174 & 177 & 197 & 150 & 153 & 193 \\
        \end{tabular}}
	\caption{FPS under the same budget and with Improved-GS as the shared pre-stage.}
    \label{tab:fps}
\end{table}

Table~\ref{tab:fps} reports FPS under the same budget and with Improved-GS as the shared pre-stage. 
Our method achieves slightly higher FPS than Improved-GS.

FPS is primarily determined by the average length of the per-pixel rendering queue. 
MiniGS significantly improves FPS by employing depth reinitialization and a ``max-contribution-only'' retention rule, both of which shorten rendering queues. 
However:
\begin{itemize}
    \item depth reinitialization incurs considerable extra optimization time, and
    \item retaining only the maximum-contribution Gaussian contradicts the objective of preserving the highest rendering quality.
\end{itemize}

Since our focus is on improving \textbf{quality} in compact 3DGS rather than maximizing FPS, direct comparison with MiniGS is not entirely appropriate. Compared with other pruning methods, our approach yields no noticeable FPS disadvantage while achieving superior quality.

\subsection{Reproduction Issue of GaussianSPA}

\begin{table}[ht]
	\centering
		\begin{tabular}{l|cccc}
			Dataset & \multicolumn{4}{c|}{Mip-NeRF360}\\
			Method|Metric
			& $SSIM^\uparrow$ & $PSNR^\uparrow$  & $LPIPS^\downarrow$  & $Num^\downarrow$\\
			\hline
			MiniGS & 0.822  & 27.36  & 0.217  & 493466\\
            GaussianSPA-30K & 0.817  & 27.31  & 0.229  & 421427\\
            GaussianSPA-40K & 0.821  & 27.58  & 0.224  & 421086\\
        \end{tabular}
	\caption{Quantitative results based on 3DGS densification method.}
	\label{tab:spa}
\end{table}

The GaussianSPA paper states that pruning is based solely on opacity. 
However, the official implementation instead ranks Gaussians using:
\[
\text{current opacity} + \text{accumulated decay offset}.
\]
In each pruning step, Gaussians within the top budget are exempted, whereas the others undergo decay. 
The accumulated offset corresponds to past decay, and due to the implemented ranking rule, Gaussians that have undergone multiple decays (i.e., originally low-opacity Gaussians) gradually move forward in the ranking, making them increasingly likely to be retained.

Thus, the implementation effectively becomes:
\[
\text{opacity pruning} + \text{partial random pruning},
\]
introducing unintended stochasticity.  
When we correct this logic, performance noticeably decreases—likely because the stochastic component implicitly mimics the gradient-competition behavior of our natural selection, thereby improving performance ``by accident.''

SPA’s true contribution lies in its ADMM-based smooth attenuation mechanism, which is independent of any specific pruning weight. 
Although a MiniGS-based pruning weight is provided, it requires computing per-view maximum contribution counts, making training several times slower. 
For fairness, we report the original SPA results in the main paper and explain the discrepancy in the appendix.

Additionally, SPA uses \textbf{40K} iterations in its paper while all comparison methods use \textbf{30K}. 
According to the authors, SPA continues to improve beyond 30K whereas other methods overfit. 
For fair comparison, we report SPA at 30K in the main paper, and its 40K results are included in Table~\ref{tab:spa}.

\begin{table}[ht]
	\centering
		\begin{tabular}{l|cccc}
			Method|Metric
			& $SSIM^\uparrow$ & $PSNR^\uparrow$  & $LPIPS^\downarrow$  & $Num^\downarrow$\\
			\hline
			base & 0.792  & 25.78  & 0.213\\
            $T 2.5\times and lr 0.4\times$ & 0.792  & 27.78  & 0.212\\
        \end{tabular}
	\caption{Different combinations of $T$ and learning rate in quality.}
	\label{tab:t}
\end{table}

\subsection{Effect of the Finite Prior Parameter T}

Table~\ref{tab:t} presents results obtained using different combinations of $T$ and learning rate.  
Increasing $T$ is theoretically equivalent to increasing the learning rate because, under the finite-prior formulation:
\[
\nabla v = 2(\mathbb{E}[v] - T).
\]
When $T \gg \mathbb{E}[v]$, the gradient magnitude is dominated by $T$, making $|\nabla v|$ proportional to $|T|$.

When $T$ is scaled to 2.5 times its original value, the effective gradient scaling factor becomes:
\[
\frac{2.5T - \mathbb{E}[v]}{T - \mathbb{E}[v]}
= 2.5 + \frac{1.5\mathbb{E}[v]}{T - \mathbb{E}[v]}.
\]
Since $\mathbb{E}[v] < 0$ in practice (as described in the main text), the second term is positive, meaning the effective growth factor is \emph{greater} than $2.5$.

\begin{figure}[t]
    \centering
    \includegraphics[width=0.8\textwidth]{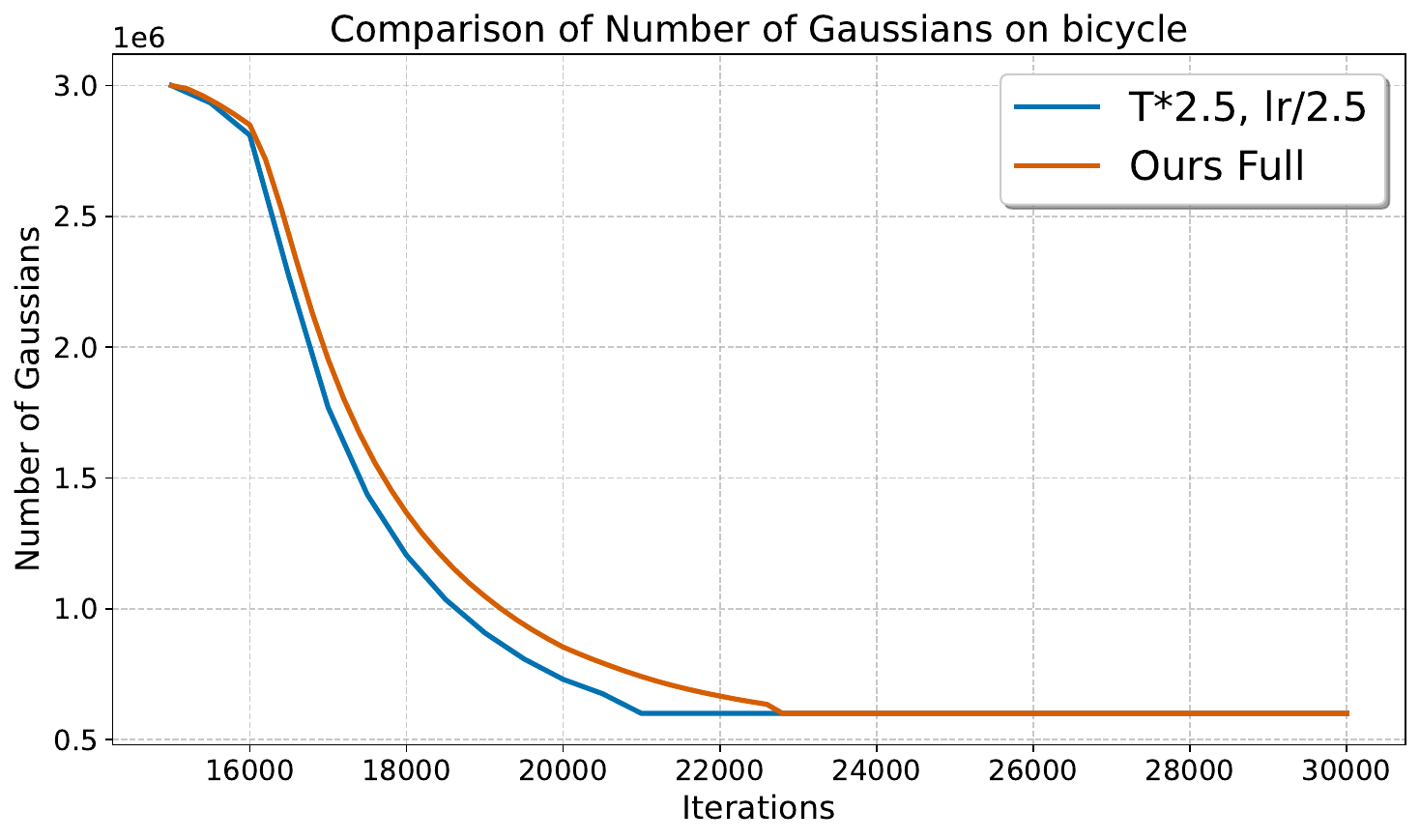}
    \caption{Different combinations of $T$ and learning rate in pruning speed.}
    \label{fig:t}
\end{figure}

Thus, if $T$ increases by $2.5\times$ while the learning rate is reduced by $2.5\times$, the actual natural-selection speed becomes slightly faster—consistent with the observed results in Fig.~\ref{fig:t}.

\end{document}